\documentclass[runningheads]{llncs}
\usepackage{graphicx}
\usepackage{tikz}
\usepackage{comment}
\usepackage{amsmath,amssymb} 
\usepackage{color}
\usepackage{cases}
\usepackage[width=122mm,left=12mm,paperwidth=146mm,height=193mm,top=12mm,paperheight=217mm]{geometry}

\usepackage[pagebackref=true,breaklinks=true,letterpaper=true,colorlinks,bookmarks=false]{hyperref}
\usepackage{multirow}
\usepackage{caption}
\usepackage{subcaption}
\graphicspath{{figures/}}
\usepackage{gensymb}
\usepackage{float}
\usepackage{bbding}
\usepackage{array}
\usepackage{booktabs}

\begin{document}
\pagestyle{headings}
\mainmatter
\def\ECCVSubNumber{618}  

\title{Progressively Guided Alternate Refinement Network for RGB-D Salient Object Detection} 

\titlerunning{Progressively Guided Alternate Refinement Network}
%
\author{Shuhan Chen\inst{1} \and Yun Fu\inst{2}}
\authorrunning{Shuhan Chen et al.}
%
\institute{School of Information Engineering, Yangzhou University, Yangzhou, China \and Department of ECE and Khoury College of Computer Science, Northeastern University, Boston, USA
\\
\email{shchen@yzu.edu.cn, yunfu@ece.neu.edu}}
\maketitle

\begin{abstract}
In this paper, we aim to develop an efficient and compact deep network for RGB-D salient object detection, where the depth image provides complementary information to boost performance in complex scenarios. Starting from a coarse initial prediction by a multi-scale residual block, we propose a progressively guided alternate refinement network to refine it. Instead of using ImageNet pre-trained backbone network, we first construct a lightweight depth stream by learning from scratch, which can extract complementary features more efficiently with less redundancy. Then, different from the existing fusion based methods, RGB and depth features are fed into proposed guided residual (GR) blocks alternately to reduce their mutual degradation. By assigning progressive guidance in the stacked GR blocks within each side-output, the false detection and missing parts can be well remedied. Extensive experiments on seven benchmark datasets demonstrate that our model outperforms existing state-of-the-art approaches by a large margin, and also shows superiority in efficiency (\textbf{71 FPS}) and model size (\textbf{64.9 MB}).
\end{abstract}

\section{Introduction}
The goal of salient object detection (SOD) is to detect and segment the objects or regions in an image or video~\cite{fan2019shifting} that visually attract human attention most. It is usually serves as a pre-processing step to benefit a lot of vision tasks, such as image-sentence matching~\cite{Ji_2019_ICCV}, weakly-supervised semantic segmentation~\cite{wei2016stc}, few-shot learning~\cite{zhang2019few}, to name a few. Benefiting from the rapid development of deep convolutional neural networks (CNNs), it has achieved profound progresses recently. Nevertheless, it is still very challenging in some complex scenes, such as low contrast, objects sharing similar appearance with its surroundings~\cite{fan2018salient}.

RGB-D cameras are now easily available with low-price and high-performance, such as RealSense, Kinect, which can provide depth image that contains necessary geometric information. Utilizing depth additional to the RGB image could potentially improve the performance in the above challenging cases, which is also proven to be an effective way in the applications of object detection~\cite{li2018cross}, semantic segmentation~\cite{lin2018scn}, and crowd counting~\cite{lian2019density}.

\begin{figure*}  
  \centering  
  \includegraphics[width=122mm]{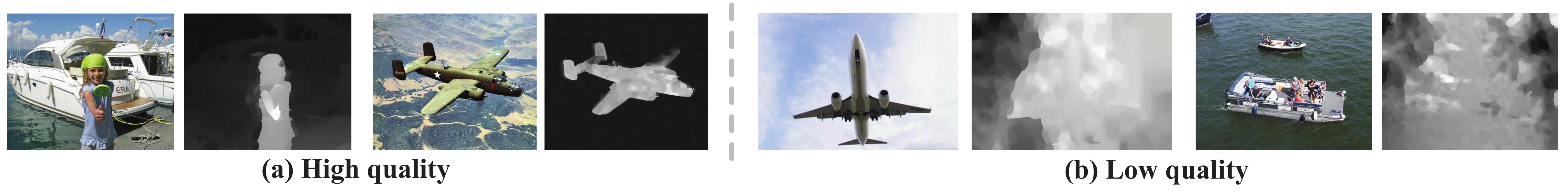}
  \caption{Example RGB images with corresponding depth images. (a) High quality depth images successfully pop out salient objects, thus can be seen as mid-level or high-level feature maps. (b) Low quality depth images are cluttered thus may be harmful for the prediction.}  
  \label{fig_depth}  
\end{figure*}

Although several novel CNN-based SOD approaches~\cite{piao2019depth}\cite{zhao2019contrast} have been proposed for RGB-D data recently, the optimal way to fuse RGB and depth information remains an open issue, which lies in two aspects: model incompatibility and redundancy, low-quality depth map. Most of the existing fusion strategies can be classified into early fusion~\cite{qu2017rgbd}\cite{song2017depth}, late fusion~\cite{han2017cnns}\cite{fan2019rethinking}, and middle fusion~\cite{chen2018progressively}\cite{chen2019multi}\cite{chen2019three}\cite{piao2019depth}. Recent researches mainly focus on the middle fusion where a separate backbone network pre-trained from ImageNet~\cite{deng2009imagenet} is usually utilized to extract depth features, which may causes incompatible  problem due to the inherent modality difference between RGB and depth image~\cite{zhao2019contrast}. Besides that, such two-stream framework doubles the number of model parameters and computation cost, which is not efficient and also contains much redundancy. Furthermore, depth maps may vary in qualities due to the limitations of the depth sensors, high-quality depth map can well pop-out salient objects with well-defined closed boundaries, while low-quality ones are cluttered and may be noisy to the prediction, as shown in Fig.~\ref{fig_depth}.

To address the above issues, we first construct a depth stream to extract depth features by learning from scratch without using pre-trained backbone network. As we know, RGB and depth have very different properties. Depth only captures the spatial structure and 3D layout of objects, without any texture details, thus contains much less information than RGB image. Since lacking low-level information, it is redundant to use pre-trained backbone network to extract features, especially these shallow layers. As seen in Fig.~\ref{fig_depth}, the high quality depth image can be seen as a mid-level or high-level feature map. Based on this observation, we design a lightweight depth stream to capture complementary high-level features only, specifically, four convolutional layers are sufficient to achieve it. Therefore, it is not only much more compact and efficient than existing two-backbone based models, but also without the incompatible problem.

Instead of directly fusing RGB features and depth features (\textit{e.g.}, concatenation or summation), which may degrade the confident RGB features especially when the input depth is noisy, we propose a alternate refinement strategy to incorporate them separately. The whole architecture of the proposed network follows a coarse-to-fine framework which starts from a initial prediction generated by a proposed Multi-Scale Residual (MSR) block. Then, it is refined progressively from deep side-outputs to shallow ones. To alleviate the information dilution in the refinement process, we propose a novel guided residual (GR) block where input prediction map is used to guide input feature to generate refined prediction and refined feature, which will be further fed into a following GR block for subsequent refinement. By assigning different guidance roles into different stacked GR blocks within each side-output, the input coarse prediction can be refined progressively into more complete and accurate.

Experimental results over 7 benchmark datasets demonstrate that our model significantly outperforms state-of-the-art approaches, and also with advantages in efficiency (\textbf{71 FPS}) and compactness (\textbf{64.9 MB}). In summary, our main contributions can be concluded as follows:
\begin{itemize}

\item We construct a lightweight depth stream to extract complementary depth features, which is much more compact and efficient than using pre-trained backbone network while without the incompatible problem.

\item We propose an alternate refinement strategy by feeding RGB feature and depth feature alternately into GR blocks, in this way to avoid breaking the good property of the confident RGB feature when the depth is low quality.

\item We further design a guided residual block to address the information dilution issue, where an input prediction is used as a guidance to generate both refined feature and refined prediction. By stacking them with progressive guidance, the missing parts and false predictions can be well remedied.
\end{itemize}

\section{Related Work}
\subsection{RGB Salient Object Detection}
\textbf{Coarse-to-Fine.} Before deep learning, salient regions are usually discovered from less ambiguous regions to difficult regions~\cite{gong2015saliency}\cite{zhang2016ranking}\cite{chen2016pr}. Following this idea, coarse-to-fine frameworks are widely explored in recent CNNs based works. Liu~\cite{liu2016dhsnet} first proposed a hierarchical recurrent convolutional neural network for refinement in a global to local and coarse to fine pipeline. Using a eye fixation map as initial prediction, Wang \textit{et al.}~\cite{wang2018salient} proposed a hierarchy of convolutional LSTMs to progressively optimize it in a top-down manner. Chen \textit{et al.}~\cite{chen2018reverse}\cite{chen2020tip} applied side-output residual learning for refinement, which is guided by a novel reverse attention block. While in~\cite{feng2019attentive}, attentive feedback module was designed for better guidance. Besides the above top-down guidance, Wang \textit{et al.}~\cite{wang2019iterative} further integrated bottom-up inference in an iterative and cooperative manner for recurrent refinement. In this paper, we also follow the coarse-to-fine pipeline, the difference is that we learn multiple residuals in each side-output with progressive guidance, which can better remedy the missing object parts and false detection.

\textbf{Top-down Guidance.} The deep layer contains high-level semantic information, which can be used as a guidance to help shallow layers filter out noisy distraction~\cite{chen2020cyb}. Such a top-down guidance manner was also widely applied in existing methods. Deep prediction maps are concatenated with shallow prediction by short connections for guidance in~\cite{hou2019deeply}, and used to erase its corresponding regions in shallow feature to guide residual learning in~\cite{chen2018reverse}. While in~\cite{wu2019cascaded}, it was utilized to weight shallow feature by the proposed holistic attention module. In~\cite{liu2019deep}, it was applied into each group of the divided shallow convolutional feature to further promote its guidance role. Liu \textit{et al.}~\cite{liu2019simple} built a global guidance module to transmit the location information into different shallow layers. Zhang \textit{et al.}~\cite{zhang2019capsal} made a further step by leveraging captioning
information to boost SOD. There are also several recent works following this effective strategy, such as~\cite{zhao2019egnet}\cite{pang2020multi}\cite{zhang2020weakly}. While in our model, the deep prediction is used as guidance both for feature refinement and prediction refinement. Furthermore, a progressive guidance strategy is proposed to address the feature dilution issue.

\subsection{RGB-D Salient Object Detection}
\textbf{Hand-crafted based.} Early works are all focusing on various hand-crafted features, including multi-contextual contrast~\cite{peng2014rgbd}, anisotropic center-surround difference~\cite{ju2014depth}, local background enclosure~\cite{feng2016local}, and so on.

\textbf{CNNs based.} To increase feature representation ability, CNNs is widely applied and has dominated this area in recent years. As mentioned above, these approaches can be roughly divided into early fusion, middle fusion, and late fusion. Early fusion regards the depth map as a additional channel to concatenate with RGB as initial input, \textit{e.g.}, \cite{song2017depth}. Late fusion applies two separate backbone network for RGB and depth to generate individual features or predictions which are fused together for final prediction, such as \cite{han2017cnns}\cite{fan2019rethinking}. Most of recent works focused on the middle fusion scheme, which incorporate multi-scale RGB features and depth features in different manners. A complementarity-aware fusion module was proposed in~\cite{chen2018progressively}. Piao \textit{et al.} \cite{piao2019depth} designed a depth refinement block to fuse multi-level paired depth and RGB cues. Instead of cross-modal feature fusion, Zhao \textit{et al.} \cite{zhao2019contrast} addressed it in a different manner by integrating enhanced depth cues to weight RGB features for better representation. In~\cite{piao2020exploit}, an asymmetrical two-stream architecture based on knowledge distillation was proposed for light field SOD. Different from them, we construct a lightweight depth stream by learning from scratch, whose depth features are fed for refinement separately.

\section{The Proposed Network}
We first present the overall architecture of the proposed alternate refinement network, and then introduce its main components in detail, including MSR block to generate coarse initial prediction, and GR block with progressive guidance. Finally, we also discuss the differences with related networks.

\begin{figure*}  
  \centering  
  \includegraphics[width=118mm]{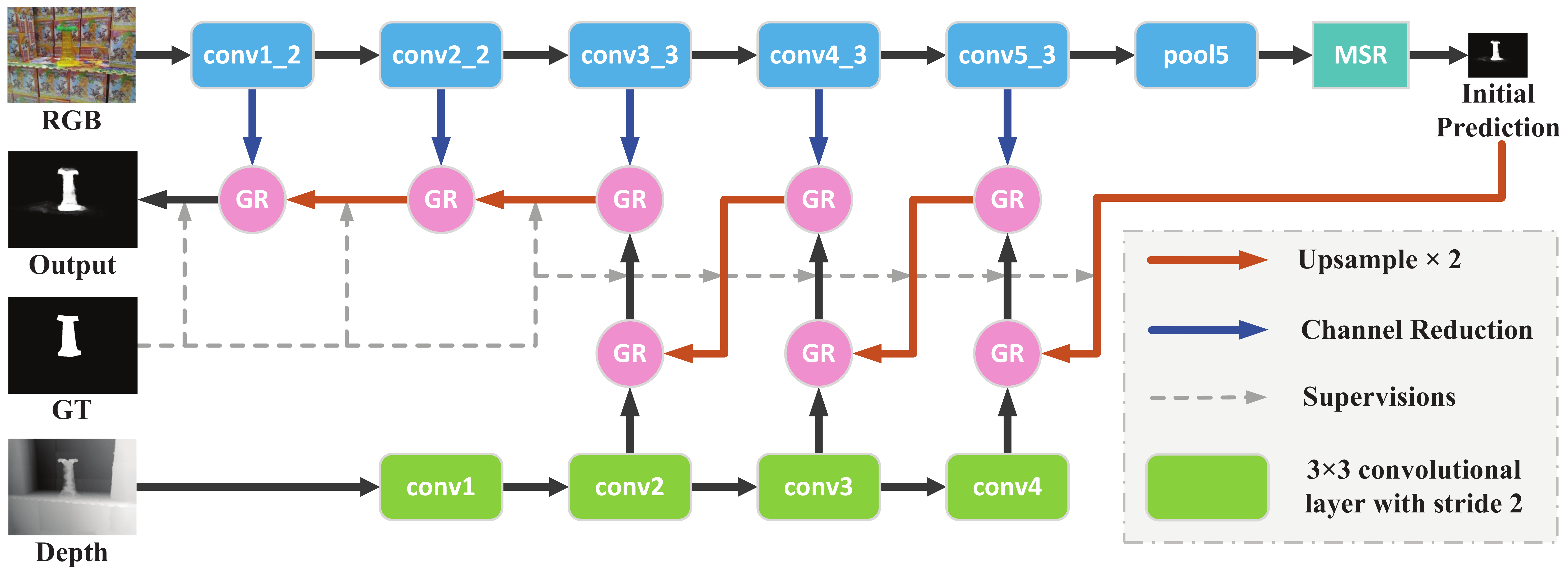}
  \caption{The overall architecture of the proposed network, where RGB feature and depth feature are fed into GR blocks \textbf{alternately} for refinement. Here, we only show single GR block in each side-output for clarity. Detailed structures of MSR and GR are illustrated in Fig.~\ref{fig_msr} and Fig.~\ref{fig_gr} respectively.}  
  \label{fig_architecture}  
\end{figure*}

\subsection{The Overall Architecture}
Our network follows the existing coarse-to-fine refinement framework as seen in Fig.~\ref{fig_architecture}. Given a coarse initial prediction generated by our MSR block, we apply our GR block to refine it progressively by combing multi-level convolutional features from RGB and depth streams alternately. Considering the modal gap between RGB and depth, furthermore, the quality of the depth varies tremendously across different scenarios due to the limitation of depth sensors, we don't directly fuse the RGB and depth features, instead, they are fed into our network alternately to reduce their mutual degradation for better refinement. Finally, we apply deep supervisions on each side-output and train the whole network in an end-to-end manner, only with standard binary cross entropy loss.

\textbf{RGB stream.} We utilize VGG16~\cite{simonyan2015very} as backbone network to extract multi-level RGB features, where \{{\texttt{conv1\_2}}, {conv2\_2}, {\texttt{conv3\_3}}, {\texttt{conv4\_3}}, {\texttt{conv5\_3}}\} are chosen as side-output features, which have \{1, 1/2, 1/4, 1/8, 1/16\} of the input image resolution respectively. We first apply $1\times1$ convolutional layers to reduce their dimensions into \{16, 32, 64, 64, 64\} for efficiency. Then, these side-output features (denoted as $F_{1}$, $F_{2}$, $F_{3}$, $F_{5}$, $F_{7}$) are used for subsequent refinement.

\textbf{Depth stream.} Instead of using pre-trained backbone network as most of the existing works did, we construct a light-weight depth stream to extract complementary features, which only consists of four cascaded $3\times3$ convolutional layers with 64 channels and stride 2. The last three layers are selected as high-level side-output features for refinement, which are denoted as $F_{4}$, $F_{6}$, $F_{8}$, with \{1/4, 1/8, 1/16\} of the input image resolution respectively.

\begin{figure*}  
  \centering  
  \includegraphics[width=118mm]{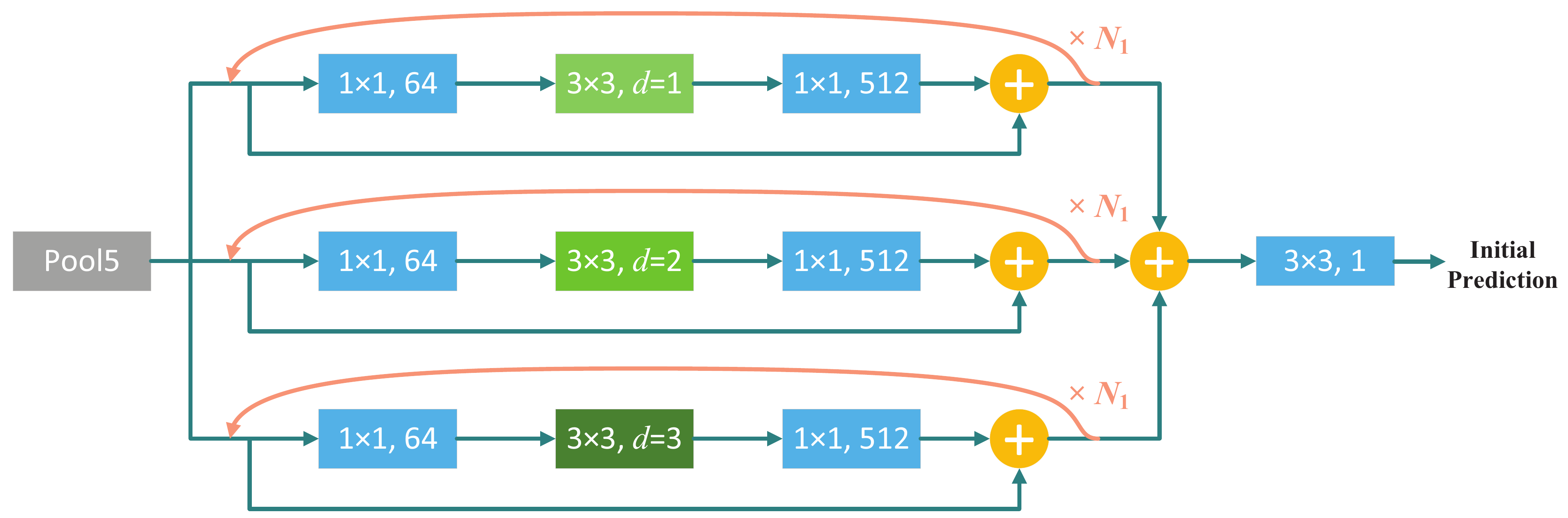}
  \caption{The proposed multi-scale residual block. ``$d$" denotes dilation rate.}  
  \label{fig_msr}  
\end{figure*}

\subsection{Multi-Scale Residual Block}
Since the scale of salient objects vary from large to small, which implies that the model needs to capture information at different contexts in order to detect objects reliably. Although the backbone network has a large enough theoretical receptive field to cover most of the large objects, the effective receptive field is smaller than the theoretical receptive field as demonstrated in~\cite{luo2016understanding}. Inspired from~\cite{li2019scale}, we design a multi-scale residual block to address the scale issue for SOD. The proposed MSR block is embedded after ``\texttt{pool5}" and consists of three parallel branches in which each shares the same residual structure except the dilation rate. Each residual block consists of three convolutions with kernel size $1\times1$, $3\times3$, and $1\times1$. The dilation rates for the $3\times3$ convolutional layers are 1, 2, and 3 respectively, as shown in Fig.~\ref{fig_msr}. Instead of stacking such residual block to increase receptive field, we implement it in a recurrent manner to reduce the number of parameters, which has been widely applied in previous works~\cite{zhang2018progressive}\cite{deng2018r3net}\cite{li2018recurrent}. The total recurrent iteration is $N_{1}$ for each branch. Finally, all the branches are added together then fed into a $3\times3$ convolutional layer to produce a single channel initial prediction.

Although shares similar structure, the proposed MSR differs from~\cite{li2019scale} in the following two aspects. Firstly, since the scale-aware ground truth is not easy to obtain in SOD, we don't share weights among different branches but different stacked blocks in each branch. Secondly, these branches are fused together for the initial prediction. 

\subsection{Guided Residual Block}
\begin{figure*}  
  \centering  
  \includegraphics[width=118mm]{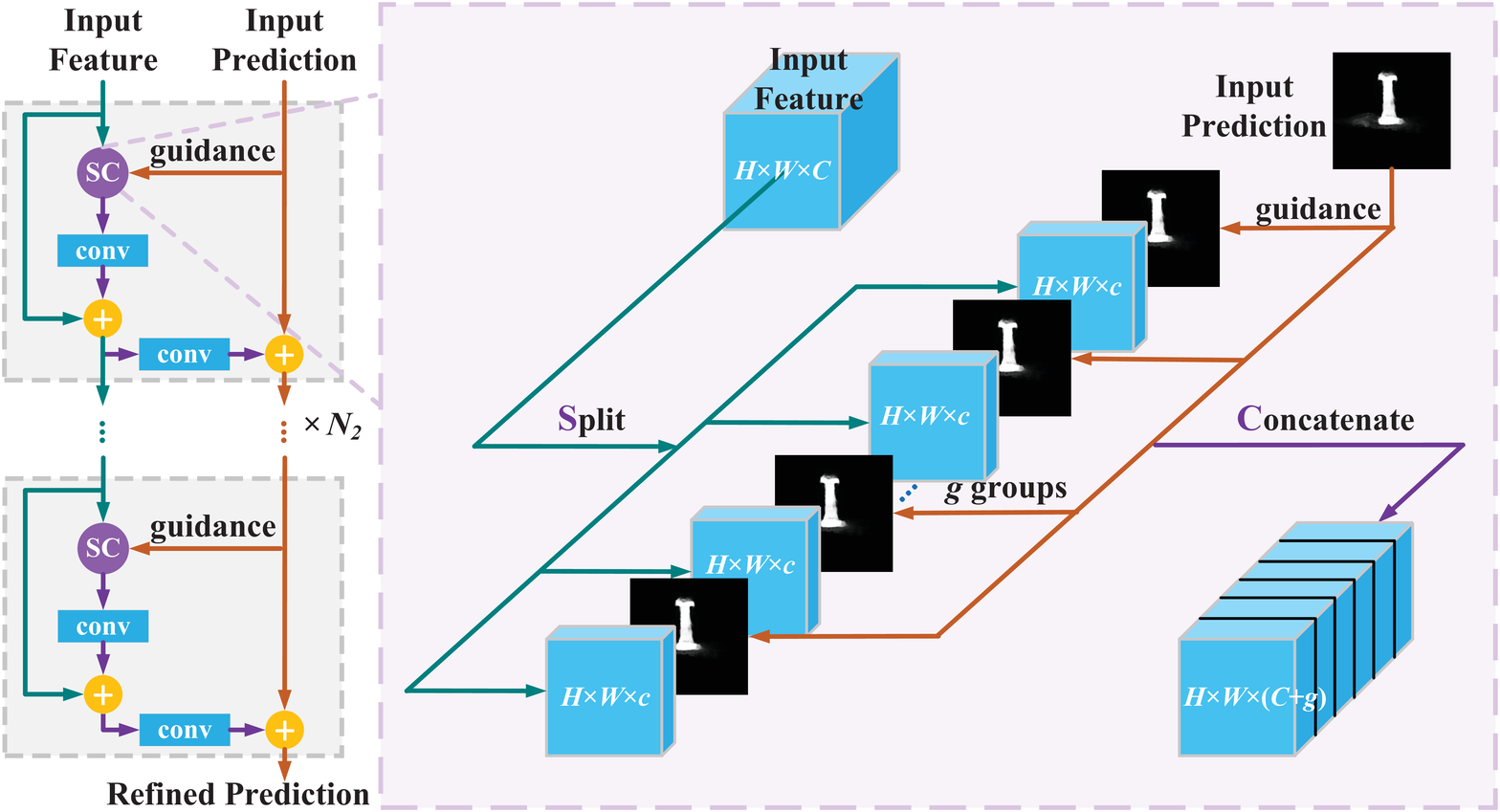}
  \caption{The proposed guided residual blocks (dashed bounding boxes on the left), which are stacked with progressive guidance. SC denotes split-and-concatenate operation (dashed bounding box on the right).}  
  \label{fig_gr}  
\end{figure*}

As we know, different layers of deep CNNs learn different scale features, shallow layers capture low-level structure cues while deep layers capture high-level semantic information. Based on this observation, various fusion strategies were proposed to combine their complementary cues, such as short connection~\cite{hou2019deeply}, skip connection~\cite{lin2017feature}, residual connection~\cite{he2016deep}\cite{deng2018r3net}\cite{zhang2020residual}. However, the high-level information of deep layer may be gradually diluted in the fusion process, especially when combing with the noisy shallow features. To address it, we design a novel guided residual block which composes of two parts: split-and-concatenate (SC) operation, and dual residual learning. As illustrated in Fig.~\ref{fig_gr}, given an input feature and prediction map, it outputs refined alternatives.

\textbf{Split-and-Concatenate.} Given the convolutional feature $F$ with $C$ channels and prediction map $S$ as inputs, we first split $F$ into $g$ groups, each of which has $c$ channels. Then $S$ is utilized as a guidance feature map to be concatenated with each split feature maps. After concatenation, we obtain a $C+g$ channel feature. Based on it, the SC operation can be formulated as:

\begin{equation}
{F^{1},...,F^{j},...,F^{g}}={\rm Split}(F), j\in{1,2,...,g},
\end{equation}

\begin{equation}
F_{cat}={\rm Cat}(F^{1}, S,...,F^{j}, S,...,F^{g}, S),
\end{equation}
where Cat denotes concatenate operation and $F_{cat}$ is the concatenated feature.

\textbf{Dual Residual Learning.} After SC operation, we feed the concatenated feature $F_{cat}$ into a $3\times3$ convolutional layer for guided learning and reducing channel number into $C$, then are added with the input feature as refined output feature. Thus, the first residual learning can be formulated as:

\begin{equation}
\hat{F}=F+{\rm Conv}(F_{cat}; \theta_{1}),
\end{equation}
where ${\rm Conv}(*;\theta)$ is the convolution operation with parameter $\theta$. Another $3\times3$ convolutional layer is further applied to produce a single channel residual prediction. Based on it, we can obtain the refined prediction map $\hat{S}$ by:

\begin{equation}
\hat{S}=S+{\rm Conv}(\hat{F}; \theta_{2}).
\end{equation}
$\hat{F}$ and $\hat{S}$ will be further fed into subsequent GR block for guided residual learning.

\subsection{Progressive Guidance}
Based on the above GR block, we propose a progressive residual refinement framework. Specifically, in each side-output, we stack $N_{2}$ GR blocks, which is set to 3 in this paper, and the first GR block takes $F^{0}_{i}$ and $S^{0}_{i}$ as input feature and prediction map:

\setlength{\tabcolsep}{6pt}
\begin{table}[]
\begin{center}
\caption{The detailed setting of different guidance styles, including uniform guidance and progressive guidance. \{*,*,*\} represents the channel number $c$ in each split group from side-output 1 to side-output 3. The rest are the same with side-output 3. GR$^{r}$ denotes the $r$th GR block in each side-output.}
\label{table_gui}
\begin{tabular}{ccccc}
\hline
\hline
\specialrule{0em}{1pt}{1pt}
Guidance Style  & No. & GR$^{1}$ & GR$^{2}$ & GR$^{3}$ \\ \hline
\multirow{4}{*}{Uniform Guidance}     & 1   & \multicolumn{3}{c}{$c$=\{16,32,64\}}                     \\
                                      & 2   & \multicolumn{3}{c}{$c$=\{8,8,8\}}                          \\
                                      & 3   & \multicolumn{3}{c}{$c$=\{4,4,4\}}                          \\
                                      & 4   & \multicolumn{3}{c}{$c$=\{1,1,1\}}                          \\ \hline
\multirow{4}{*}{Progressive Guidance} & 5   & $c$=\{16,32,64\} & $c$=\{8,8,8\} & $c$=\{4,4,4\} \\
                                      & 6   & $c$=\{16,32,64\} & $c$=\{8,8,8\} & $c$=\{1,1,1\} \\
                                      & 7   & $c$=\{16,32,64\}      & $c$=\{4,4,4\} & $c$=\{1,1,1\} \\
                                      & 8   & $c$=\{8,8,8\}      & $c$=\{4,4,4\} & $c$=\{1,1,1\} \\ \hline
\hline 
\end{tabular}
\end{center}
\end{table}
\setlength{\tabcolsep}{1.4pt}

\begin{numcases}{}    S^{0}_{i}={\rm Up}(S_{i+1},F_{i}), \ {\rm if} \ i=1,2,4,6,8;\\    S^{0}_{i}=S_{i+1}, \ {\rm if} \ i=3,5,7;  \end{numcases}
in which Up$(x,y)$ represents bilinear interpolation operation that upsamples $x$ to the same size as $y$, and $i$ denotes side-output stage. Then the output of the first GR block can be denoted as $\hat{F}^{1}_{i}$ and $\hat{S}^{1}_{i}$, which will be fed into the following GR block. The last GR block only outputs the refined prediction $S_{i}=\hat{S}^{N_{2}}_{i}$ as shown in Fig.~\ref{fig_gr}. $S_{1}$ is fed into a sigmoid layer as final output.

The channel number $c$ in each split group or group number $g$ is essential for guidance. We can define different guidance styles by varying $c$ or $g$. If $g=1$, as~\cite{deng2018r3net} did, the guidance role is very weak due to the imbalanced channels ($C$ versus 1). In~\cite{liu2019deep}, $c$ is set to 4 in all the side-outputs, which can be seen as a medium guidance. Extremely, the guidance role is very strong when we set $c$ to 1. Different from them, we first define uniform guidance by sharing the same guidance role in all the stacked blocks, as listed in Table~\ref{table_gui}. Since the prediction map will becomes more and more accurate in the refinement process, we further define progressive guidance by gradually increasing the guidance role. We will conduct ablation experiments to investigate the best setting in Section~\ref{sec4.2}.

Such a progressive residual refinement inherits the following good properties. The stacked residual units establish multiple shortcut connections between each side-output prediction and the ground truth, which enables it easier to remedy the missing object parts and detection errors. Extremely, with the strong supervision on each side-output, the error is approximately equal to zero if there is no useful information in the input feature, \textit{e.g.}, when the depth image is low quality. In this way, we can greatly reduce its noisy distraction, thus leads to more accurate detection. Furthermore, such residual units also enhance the input feature gradually for better refinement.

\subsection{Difference to Other Networks}
Although shares the same split-and-concatenate operation, the proposed GR block differs from the group guidance module (GGM)~\cite{liu2019deep} in two aspects. (1) GGM apply group convolution on the output of SC, which only focuses on each split group for guidance. Different from it, the convolution is performed on all the concatenated feature maps in our GR, which benefits the information passing among different groups. (2) The channel number in each split group is fixed to 4 in GGM. While our GR blocks are stacked with progressive guidance by varying different $c$. Our network also differs other residual learning based architectures, \textit{e.g.}, RAS~\cite{chen2018reverse}\cite{chen2020tip}, R$^{3}$Net~\cite{deng2018r3net}. Firstly, we learn dual residuals progressively in each side-output which can better remedy the missing object parts and false detection in the initial prediction. Secondly, the prediction maps are progressively applied for guidance during residual refinement. The effectiveness and superiority will be verified in the following section.

\section{Experimental Results}
\subsection{Experimental Setup}
\textbf{Datasets.}
We adopt 7 widely used RGB-D benchmark datasets for evaluation, including NJUD~\cite{ju2014depth}, NLPR~\cite{peng2014rgbd}, DES~\cite{cheng2014depth}, STERE~\cite{niu2012leveraging}, LFSD~\cite{li2014saliency},  DUT~\cite{piao2019depth}, and SIP~\cite{fan2019rethinking}, which contain 1985, 1000, 135, 1000, 100, 1200, 929 well annotated images, respectively. Among them, SIP is a recent collected human activities oriented dataset with high image resolution (744$\times$992). To make a fair comparison, we follow the same training settings as existing works~\cite{han2017cnns}\cite{chen2018progressively}\cite{chen2019three}\cite{fan2019rethinking}, which consists of 1485 samples from NJUD and 700 samples from NLPR. To reduce over-fitting risk, we augment the input image by random horizontal flipping and rotating (0\degree, 90\degree, 180\degree, 270\degree), which increases the training images by four times.

\textbf{Evaluation Metrics.}
We adopt five widely applied metrics for comprehensive evaluation, \textit{i.e.}, precision-recall (PR) curve, F-measure ($F_{\beta}$), S-measure~\cite{fan2017structure} ($S_{\alpha}$), E-measure~\cite{fan2018enhanced} ($E_{\xi}$), and mean absolute error ($M$). Specifically, PR curve is plotted via pairs of precision and recall values which are calculated by comparing the binary saliency map with its ground truth. $F_{\beta}$ is an overall metric and only its maximum value is reported here, where $\beta^{2}$ is set to 0.3 to emphasize the precision over recall. $S_{\alpha}$ and $E_{\xi}$ are two recent proposed metrics which evaluate the spatial structure similarities, local pixel matching and image-level statistics information, respectively. Higher scores of $E_{\xi}$, $S_{\alpha}$, and $F_{\beta}$ indicate better performance, while lower for $M$,.

\textbf{Implementation Details.}
We implemented our method in PyTorch~\cite{paszke2019pytorch} and on a PC with single NVIDIA TITAN Xp GPU. All the images are resized to 352$\times$352 both for training and inferring. The depth image needs to be normalized into [0, 1]. The proposed model is trained by Adam optimizer~\cite{kingma2015adam} with the following hyper-parameters: batch size (10), epochs (30), initial learning rate (1e-4), which is decreased by 10 after 25 epochs. The parameters of the backbone network in the RGB stream is initialized by VGG16~\cite{simonyan2015very}, while the others are using the default setting of the Pytorch. We will release the source code for research purpose on \url{http://shuhanchen.net}.

\setlength{\tabcolsep}{6pt}
\begin{table}[]
\begin{center}
\caption{Quantitative comparison of different settings in Table~\ref{table_gui}.}
\label{table_gs}
\begin{tabular}{ccccccccc}
\hline
\hline
No.  & 1   & 2   & 3   & 4   & 5   & 6   & 7 & 8  \\ \hline
$E_{\xi}\uparrow$ & 0.903 & 0.903 & 0.903 & 0.904 & 0.903 & \textbf{0.908} & 0.907 & 0.905 \\
$S_{\alpha}\uparrow$ & 0.866 & 0.869 & 0.869 & 0.870 & 0.867 & \textbf{0.875} & 0.871 & 0.871 \\
$F_{\beta}\uparrow$ & 0.845 & 0.845 & 0.845 & 0.846 & 0.845 & \textbf{0.848} & 0.847 & 0.842 \\
$M\,\downarrow$ & 0.062 & 0.062 & 0.062 & 0.060 & 0.061 & 0.059 & \textbf{0.058} & 0.059 \\ \hline
\hline
\end{tabular}
\end{center}
\end{table}
\setlength{\tabcolsep}{1.6pt}

\setlength{\tabcolsep}{6pt}
\begin{table}[]
\begin{center}
\caption{Quantitative comparison with different ablation settings. R and D denote RGB stream and depth stream respectively. St: stacking 7 MSR blocks; Re: proposed recurrent strategy. MS: model size (MB).}
\label{table_ablation}
\begin{tabular}{r|cc|cc|cc|cc}
\hline
\hline
          & St    & Re    & Cat   & AR    & R     & R+D   & VGG16 & Ours  \\ \hline
$E_{\xi}\uparrow$ & \textbf{0.907} & \textbf{0.907} & 0.899 & \textbf{0.908} & 0.886 & \textbf{0.908} & 0.896 & \textbf{0.908} \\
$S_{\alpha}\uparrow$ & 0.871 & \textbf{0.872} & 0.863 & \textbf{0.875} & 0.846 & \textbf{0.875} & 0.857 & \textbf{0.875} \\
$F_{\beta}\uparrow$ & 0.850 & \textbf{0.851} & 0.840 & \textbf{0.848} & 0.814 & \textbf{0.848} & 0.837 & \textbf{0.848} \\
$M\,\downarrow$  & \textbf{0.059} & \textbf{0.059} & 0.064 & \textbf{0.059} & 0.072 & \textbf{0.059} & 0.065 & \textbf{0.059} \\
MS$~\downarrow$ & 72.3  & \textbf{64.9}  & \textbf{63.0}  & 64.9  & \textbf{62.5} & 64.9  & 123.6 & \textbf{64.9}  \\ \hline
\hline
\end{tabular}
\end{center}
\end{table}
\setlength{\tabcolsep}{1.6pt}

\subsection{Ablation Analyses}\label{sec4.2}
We first investigate different design options and the effectiveness of different components in the proposed network on a recent challenging dataset SIP~\cite{fan2019rethinking}.

\textbf{Recurrent strategy.} To verify the effectiveness of the recurrent strategy in MSR, we first made an experiment by comparing with stacking 7 blocks with 7 iterations. Here we adopt the No. 1 setting in Table~\ref{table_gui} as guidance style. As can be seen in Table~\ref{table_ablation}, their performance are almost the same, but the recurrent strategy achieved more compact model size. We further conduct ablation study to explore how many recurrent iterations are needed in MSR by varying it $N_{1}$ from 1 to 7. The results in Fig.~\ref{fig_pr}(f) show that when $N_{1}$ grows beyond 5, the performance becomes stable. Therefore, for efficiency, we adopt the recurrent strategy and set $N_{1}$ to 5 in the following experiments.

\textbf{Guidance style.} To investigate the best setting of the guidance style, we compare the performance of all the listed settings in Table~\ref{table_gui}. From the results in Table~\ref{table_gs}, we can observe that progressive guidance shows better performance than uniform guidance, which supports our claim that the guidance role should be strengthened with the progressively refined prediction map. The No.6 setting that achieved best performance was adopt as our final guidance strategy.

\textbf{Alternate refinement.} We further conduct experiment to verify the proposed alternate refinement (AR) strategy by comparing with directly concatenating (Cat) the RGB and depth features. As shown in Table~\ref{table_ablation}, our proposed AR strategy performs better than Cat, which demonstrates our analysis that Cat may break the good property of the RGB features.

\setlength{\tabcolsep}{6pt}
\begin{table}[]
\begin{center}
\caption{Quantitative comparison with different side-output depth features.}
\label{table_dep}
\begin{tabular}{cccc|cccc}
\hline
\hline
\texttt{conv4} & \texttt{conv3} & \texttt{conv2} & \texttt{conv1} & $E_{\xi}\uparrow$ & $S_{\alpha}\uparrow$ & $F_{\beta}\uparrow$ & $M\,\downarrow$ \\ \hline
\Checkmark & &  &  & 0.892 & 0.855 & 0.828 & 0.067 \\
\Checkmark & \Checkmark & &  & 0.897 & 0.860 & 0.834 & 0.066 \\
\Checkmark & \Checkmark & \Checkmark &  & \textbf{0.908} & \textbf{0.875} & \textbf{0.848} & \textbf{0.059} \\
\Checkmark & \Checkmark & \Checkmark & \Checkmark & 0.907 & 0.873 & \textbf{0.848} & \textbf{0.059} \\ \hline
\hline
\end{tabular}
\end{center}
\end{table}
\setlength{\tabcolsep}{1.6pt}

\textbf{Depth stream.} We first evaluate the effectiveness of the constructed depth stream by removing it for comparison. As seen in Table~\ref{table_ablation}, the performance can be greatly improved when combining the depth stream, which indicates the good ability to capture complementary information by our constructed depth stream. It is also worth to note that the performance is still comparable with state-of-the-art model when only using RGB stream, which further confirms the effectiveness of our progressive residual refinement framework.

Since most of the previous works using two pre-trained backbone networks to extract RGB and depth features respectively, we also made another experiment by replacing the proposed depth stream with VGG16~\cite{simonyan2015very}. As a result, the model size and training time (4 hours) are dramatically increased with the decrease of the quantitative performance as shown in Table~\ref{table_ablation}. Therefore, our proposed depth stream is a better choice in extracting complementary depth features.

\begin{table}[t]
\scriptsize
\begin{center}
\caption{Quantitative comparison including $E_{\xi}$, $S_{\alpha}$, $F_{\beta}$, and $M$, over seven widely evaluated datasets. $\uparrow \& \downarrow$ represent higher and lower is better, respectively. $*$ denotes the models are trained on NJUD~\cite{ju2014depth}+NLPR~\cite{peng2014rgbd}+DUT~\cite{piao2019depth}, the rest are trained on NJUD~\cite{ju2014depth}+NLPR~\cite{peng2014rgbd}. The best three scores are highlighted in {\color{red} \textbf{red}}, {\color{blue} \textbf{blue}}, and \textcolor[rgb]{0.13,0.55,0.13}{\textbf{green}} respectively.}
\label{table:qua}
\begin{tabular}{p{0.7cm}<{\centering}|c|p{0.75cm}<{\centering}cp{0.75cm}<{\centering}|p{0.75cm}<{\centering}cccccp{0.75cm}<{\centering}|cc}
\hline
\hline
\multirow{2}{*}{} & \multirow{2}{*}{Metric} & LHM   & ACSD  & LBE   & DF    & CTMF  & MMCI  & TAN   & PCAN   & CPFP  & Ours  & DMRA  & Ours  \\
                         &  & \cite{peng2014rgbd} & \cite{ju2014depth}   & \cite{feng2016local}  & \cite{qu2017rgbd}     & \cite{han2017cnns}    & \cite{chen2019multi}  & \cite{chen2019three} & \cite{chen2018progressively}  & \cite{zhao2019contrast} &  & \cite{piao2019depth}$*$    & $*$  \\ \hline
\multirow{4}{*}{\rotatebox{90}{NJUD}~\rotatebox{90}{~\cite{ju2014depth}}}    & $E_{\xi}\uparrow$ & .711 & .790 & .796 & .839 & .864 & .878 & .893 & .896 & .895 & {\textcolor{blue} {\textbf{.914}}} & {\textcolor[rgb]{0.13,0.55,0.13} {\textbf{.908}}} & {\textcolor{red} {\textbf{.916}}} \\
                         & $S_{\alpha}\uparrow$ & .522 & .703 & .700 & .768 & .849 & .859 & .878 & .877 & .878 & {\textcolor{blue} {\textbf{.906}}} & {\textcolor[rgb]{0.13,0.55,0.13} {\textbf{.886}}} & {\textcolor{red} {\textbf{.909}}} \\
                         & $F_{\beta}\uparrow$ & .636 & .695 & .734 & .783 & .788 & .813 & .844 & .844 & .837 & {\textcolor{blue} {\textbf{.883}}} & {\textcolor[rgb]{0.13,0.55,0.13} {\textbf{.872}}} & {\textcolor{red} {\textbf{.893}}} \\
                         & $M\,\downarrow$ & .199 & .198 & .149 & .136 & .085 & .079 & .060 & .059 & .053 & {\textcolor{blue} {\textbf{.045}}} & {\textcolor[rgb]{0.13,0.55,0.13} {\textbf{.051}}} & {\textcolor{red} {\textbf{.042}}} \\ \hline
\multirow{4}{*}{\rotatebox{90}{NLPR}~\rotatebox{90}{~\cite{peng2014rgbd}}}    & $E_{\xi}\uparrow$ & .819 & .752 & .868 & .884 & .869 & .872 & .916 & .916 & {\textcolor[rgb]{0.13,0.55,0.13} {\textbf{.924}}} & {\textcolor{blue} {\textbf{.948}}} & {\textcolor[rgb]{0.13,0.55,0.13} {\textbf{.941}}} & {\textcolor{red} {\textbf{.955}}} \\
                         & $S_{\alpha}\uparrow$ & .631 & .684 & .777 & .806 & .860 & .856 & .886 & .874 & .888 & {\textcolor{blue} {\textbf{.918}}} & {\textcolor[rgb]{0.13,0.55,0.13} {\textbf{.899}}} & {\textcolor{red} {\textbf{.930}}} \\
                         & $F_{\beta}\uparrow$ & .665 & .548 & .747 & .759 & .723 & .730 & .796 & .795 & .822 & {\textcolor{blue} {\textbf{.871}}} & {\textcolor[rgb]{0.13,0.55,0.13} {\textbf{.854}}} & {\textcolor{red} {\textbf{.885}}} \\
                         & $M\,\downarrow$ & .103 & .171 & .073 & .079 & .056 & .059 & .041 & .044 & .036 & {\textcolor{blue} {\textbf{.028}}} & {\textcolor[rgb]{0.13,0.55,0.13} {\textbf{.031}}} & {\textcolor{red} {\textbf{.024}}} \\ \hline
\multirow{4}{*}{\rotatebox{90}{DES}~\rotatebox{90}{~\cite{cheng2014depth}}}    & $E_{\xi}\uparrow$ & .761 & .855 & .911 & .877 & .911 & .904 & .919 & .912 & .927 & {\textcolor[rgb]{0.13,0.55,0.13} {\textbf{.935}}} & {\textcolor{red} {\textbf{.944}}} & {\textcolor{blue} {\textbf{.939}}} \\
                         & $S_{\alpha}\uparrow$ & .578 & .728 & .703 & .752 & .863 & .848 & .858 & .842 & .872 & {\textcolor[rgb]{0.13,0.55,0.13} {\textbf{.894}}} & {\textcolor{blue} {\textbf{.900}}} & {\textcolor{red} {\textbf{.913}}} \\
                         & $F_{\beta}\uparrow$ & .631 & .717 & .796 & .753 & .778 & .762 & .795 & .782 & .829 & {\textcolor{blue}{\textbf{.870}}} & {\textcolor[rgb]{0.13,0.55,0.13}{\textbf{.866}}} & {\textcolor{red} {\textbf{.880}}} \\
                         & $M\,\downarrow$ & .114 & .169 & .208 & .093 & .055 & .065 & .046 & .049 & .038 & {\textcolor[rgb]{0.13,0.55,0.13} {\textbf{.032}}} & {\textcolor{blue} {\textbf{.030}}} & {\textcolor{red} {\textbf{.026}}} \\ \hline
\multirow{4}{*}{\rotatebox{90}{STERE}~\rotatebox{90}{~~\cite{niu2012leveraging}}}    & $E_{\xi}\uparrow$ & .770 & .793 & .749 & .838 & .864 & .901 & .906 & .897 & .903 & {\textcolor[rgb]{0.13,0.55,0.13} {\textbf{.917}}} & {\textcolor{red} {\textbf{.920}}} & {\textcolor{blue} {\textbf{.919}}} \\
                         & $S_{\alpha}\uparrow$ & .562 & .692 & .660 & .757 & .848 & .873 & .871 & .875 & .879 & {\textcolor{blue} {\textbf{.903}}} & {\textcolor[rgb]{0.13,0.55,0.13} {\textbf{.886}}} & {\textcolor{red} {\textbf{.907}}} \\
                         & $F_{\beta}\uparrow$ & .703 & .661 & .595 & .742 & .771 & .829 & .835 & .826 & .830 & {\textcolor{blue} {\textbf{.872}}} & {\textcolor[rgb]{0.13,0.55,0.13} {\textbf{.867}}} & {\textcolor{red} {\textbf{.880}}} \\
                         & $M\,\downarrow$ & .172 & .200 & .250 & .141 & .086 & .068 & .060 & .064 & .051 & {\textcolor{blue} {\textbf{.044}}} & {\textcolor[rgb]{0.13,0.55,0.13} {\textbf{.047}}} & {\textcolor{red} {\textbf{.041}}} \\ \hline
\multirow{4}{*}{\rotatebox{90}{SIP}~\rotatebox{90}{~\cite{fan2019rethinking}}}    & $E_{\xi}\uparrow$ & .719 & .827 & .841 & .794 & .824 & .886 & {\textcolor[rgb]{0.13,0.55,0.13} {\textbf{.893}}} & {\textcolor{blue} {\textbf{.899}}} & {\textcolor{blue} {\textbf{.899}}} & {\textcolor{red} {\textbf{.908}}} & .863 & {\textcolor{red} {\textbf{.908}}} \\
                         & $S_{\alpha}\uparrow$ & .511 & .732 & .727 & .653 & .716 & .833 & .835 & .842 & {\textcolor[rgb]{0.13,0.55,0.13} {\textbf{.850}}} & {\textcolor{blue} {\textbf{.875}}} & .806 & {\textcolor{red} {\textbf{.876}}} \\
                         & $F_{\beta}\uparrow$ & .592 & .727 & .733 & .673 & .684 & .795 & .809 & {\textcolor[rgb]{0.13,0.55,0.13} {\textbf{.825}}} & .819 & {\textcolor{blue} {\textbf{.848}}} & .819 & {\textcolor{red} {\textbf{.854}}} \\
                         & $M\,\downarrow$ & .184 & .172 & .200 & .185 & .139 & .086 & .075 & .071 & {\textcolor[rgb]{0.13,0.55,0.13} {\textbf{.064}}} & {\textcolor{blue} {\textbf{.059}}} & .085 & {\textcolor{red} {\textbf{.055}}} \\ \hline
\multirow{4}{*}{\rotatebox{90}{DUT}~\rotatebox{90}{~\cite{piao2019depth}}}    & $E_{\xi}\uparrow$ & .756 & .814 & .785 & .848 & .884 & .855 & .866 & .858 & .815 & {\textcolor[rgb]{0.13,0.55,0.13}{\textbf{.888}}} & {\textcolor{blue} {\textbf{.927}}} & {\textcolor{red} {\textbf{.944}}} \\
                         & $S_{\alpha}\uparrow$ & .551 & .706 & .679 & .733 & .834 & .791 & .808 & .801 & .749 & {\textcolor[rgb]{0.13,0.55,0.13}{\textbf{.849}}} & {\textcolor{blue} {\textbf{.889}}} & {\textcolor{red} {\textbf{.920}}} \\
                         & $F_{\beta}\uparrow$ & .683 & .699 & .668 & .764 & .792 & .753 & .779 & .760 & .736 & {\textcolor[rgb]{0.13,0.55,0.13}{\textbf{.829}}} & {\textcolor{blue} {\textbf{.884}}} & {\textcolor{red} {\textbf{.914}}} \\
                         & $M\,\downarrow$ & .179 & .181 & .236 & .144 & .097 & .113 & .093 & .100 & .100 & {\textcolor[rgb]{0.13,0.55,0.13}{\textbf{.069}}} & {\textcolor{blue} {\textbf{.048}}} & {\textcolor{red} {\textbf{.035}}} \\ \hline
\multirow{4}{*}{\rotatebox{90}{LFSD}~\rotatebox{90}{~\cite{li2014saliency}}}    & $E_{\xi}\uparrow$ & .736 & .801 & .770 & .844 & .851 & .840 & .845 & .842 & .867 & {\textcolor[rgb]{0.13,0.55,0.13}{\textbf{.869}}} & {\textcolor{red} {\textbf{.899}}} & {\textcolor{blue} {\textbf{.889}}} \\
                         & $S_{\alpha}\uparrow$ & .557 & .734 & .736 & .791 & .796 & .787 & .801 & .794 & .828 & {\textcolor[rgb]{0.13,0.55,0.13}{\textbf{.833}}} & {\textcolor{blue} {\textbf{.847}}} & {\textcolor{red} {\textbf{.853}}} \\
                         & $F_{\beta}\uparrow$ & .718 & .755 & .708 & .806 & .782 & .779 & .794 & .792 & .813 & {\textcolor[rgb]{0.13,0.55,0.13}{\textbf{.830}}} & {\textcolor{blue} {\textbf{.849}}} & {\textcolor{red} {\textbf{.852}}} \\
                         & $M\,\downarrow$ & .211 & .188 & .208 & .138 & .119 & .132 & .111 & .112 & {\textcolor[rgb]{0.13,0.55,0.13}{\textbf{.088}}} & .093 & {\textcolor{blue} {\textbf{.075}}} & {\textcolor{red} {\textbf{.074}}} \\ \hline
\hline
\end{tabular}
\end{center}
\end{table}

Finally, to investigate how many depth features are sufficient for the proposed network, we separately evaluate the performance by combing different side-output depth features. We can clearly observe from Table~\ref{table_dep} that the performance is gradually improved with the incorporation of more side-output depth features until ``\texttt{conv2}". Further incorporating ``\texttt{conv1}" doesn't bring performance gain, which supports our claim that depth image can be seen as a mid-level or high-level feature map, therefore, there is no need to explore low-level features from it with additional convolutional layers. Therefore, three side-output depth features are sufficient to capture the complementary cues.

\subsection{Comparison with State-of-the-arts}
We compare our model with 10 state-of-the-arts, consisting of 3 traditional methods: LHM~\cite{peng2014rgbd}, ACSD~\cite{ju2014depth}, LBE~\cite{feng2016local}; and 7 CNNs-based methods: DF~\cite{qu2017rgbd}, CTMF~\cite{han2017cnns}, MMCI~\cite{chen2019multi}, TAN~\cite{chen2019three}, PCAN~\cite{chen2018progressively}, CPFP~\cite{zhao2019contrast}, DMRA~\cite{piao2019depth}. Note that all the results of the compared approaches are reproduced by running source codes or pre-computed by the authors. In addition, we also trained a model (marked with $*$) using the same trainset with DMRA~\cite{piao2019depth} for fair comparison. 

\begin{figure}[t]
\captionsetup[subfigure]{labelformat=empty}	
	\centering
    \begin{subfigure}[t]{4.0cm}
		\centering
		\includegraphics[width=4.0cm]{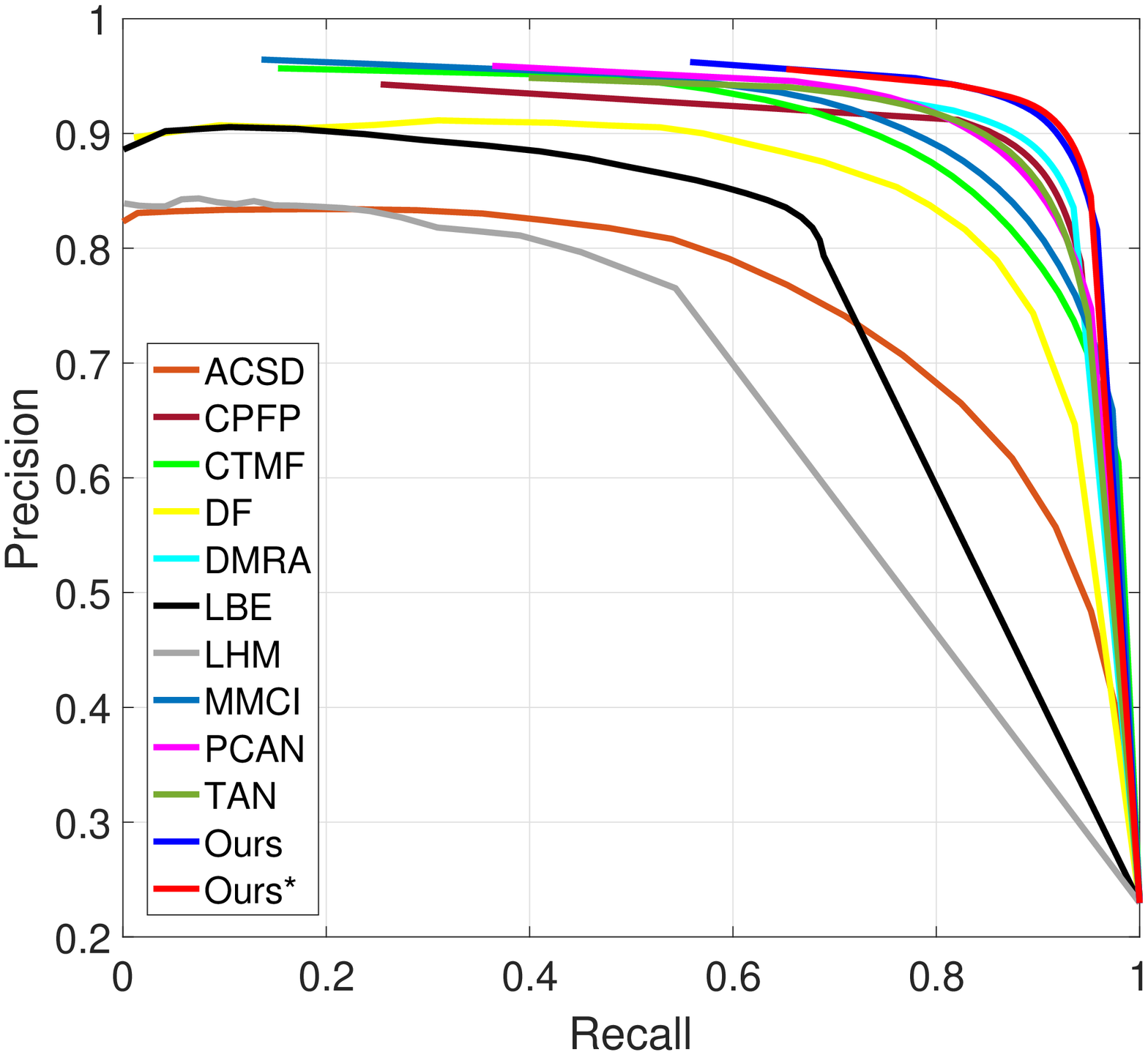}
		\caption{\scriptsize (a) NJUD~\cite{ju2014depth}}	
	\end{subfigure}
	\begin{subfigure}[t]{4.0cm}
		\centering
		\includegraphics[width=4.0cm]{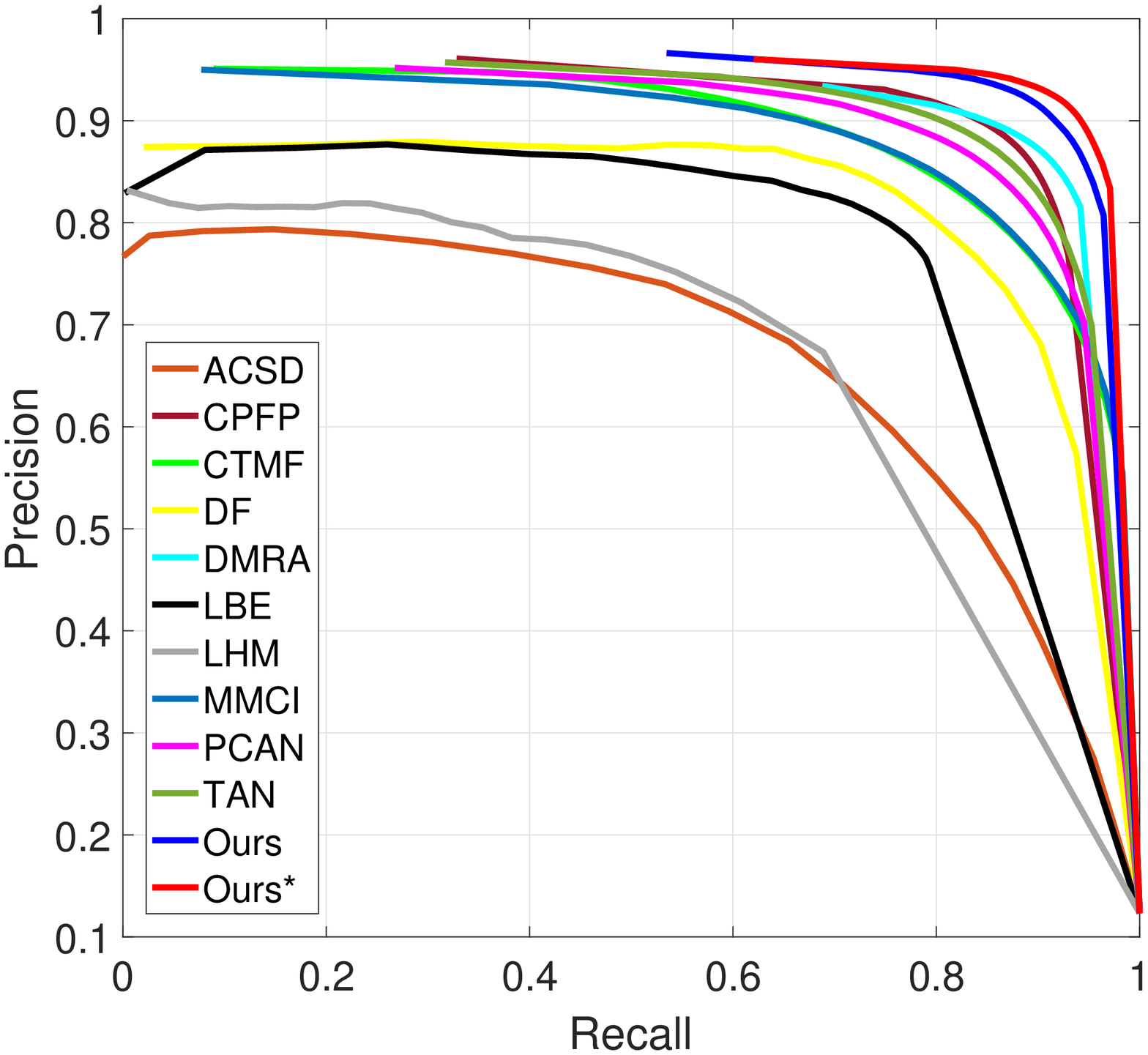}
		\caption{\scriptsize (b) NLPR~\cite{peng2014rgbd}}	
	\end{subfigure}
	\begin{subfigure}[t]{4.0cm}
		\centering
		\includegraphics[width=4.0cm]{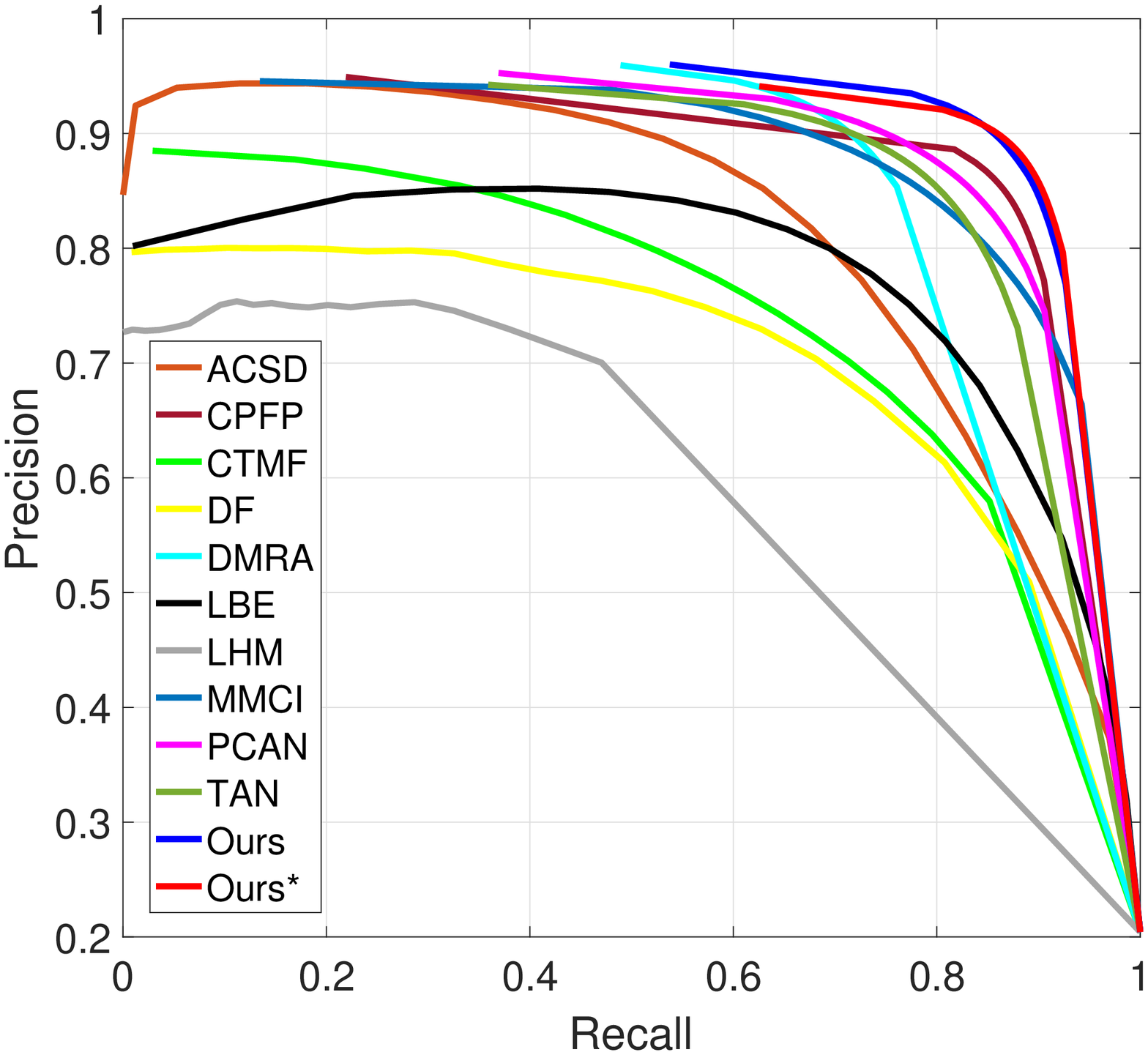}
		\caption{\scriptsize (c) SIP~\cite{fan2019rethinking}}
	\end{subfigure}

    \vspace{1pt}
    
    \begin{subfigure}[t]{4.0cm}
		\centering
		\includegraphics[width=4.0cm]{njud.eps}
		\caption{\scriptsize (d) STERE~\cite{niu2012leveraging}}	
	\end{subfigure}
	\begin{subfigure}[t]{4.0cm}
		\centering
		\includegraphics[width=4.0cm]{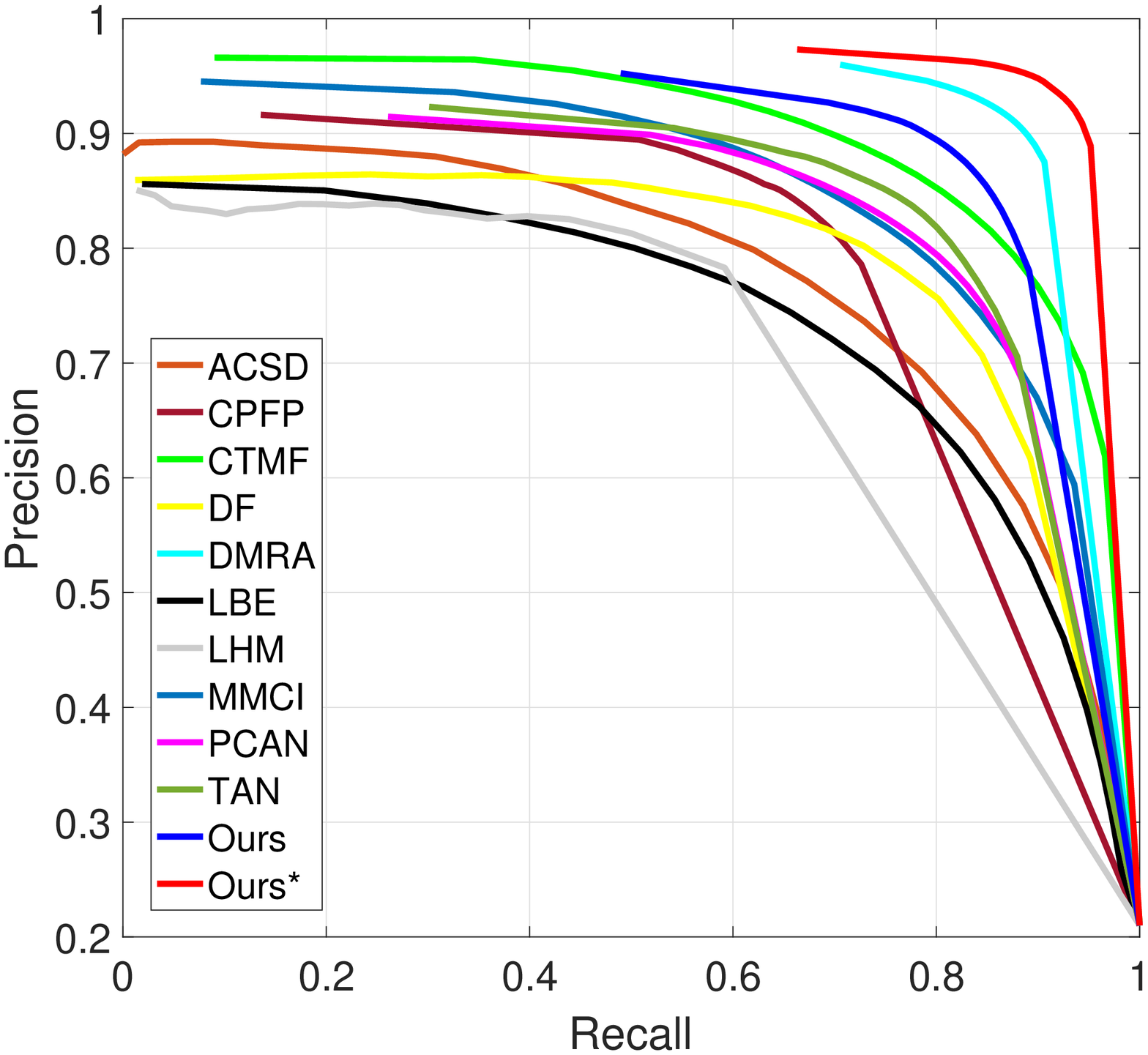}
		\caption{\scriptsize (e) DUT~\cite{piao2019depth}}	
	\end{subfigure}
	\begin{subfigure}[t]{4.0cm}
		\centering
		\includegraphics[width=4.0cm]{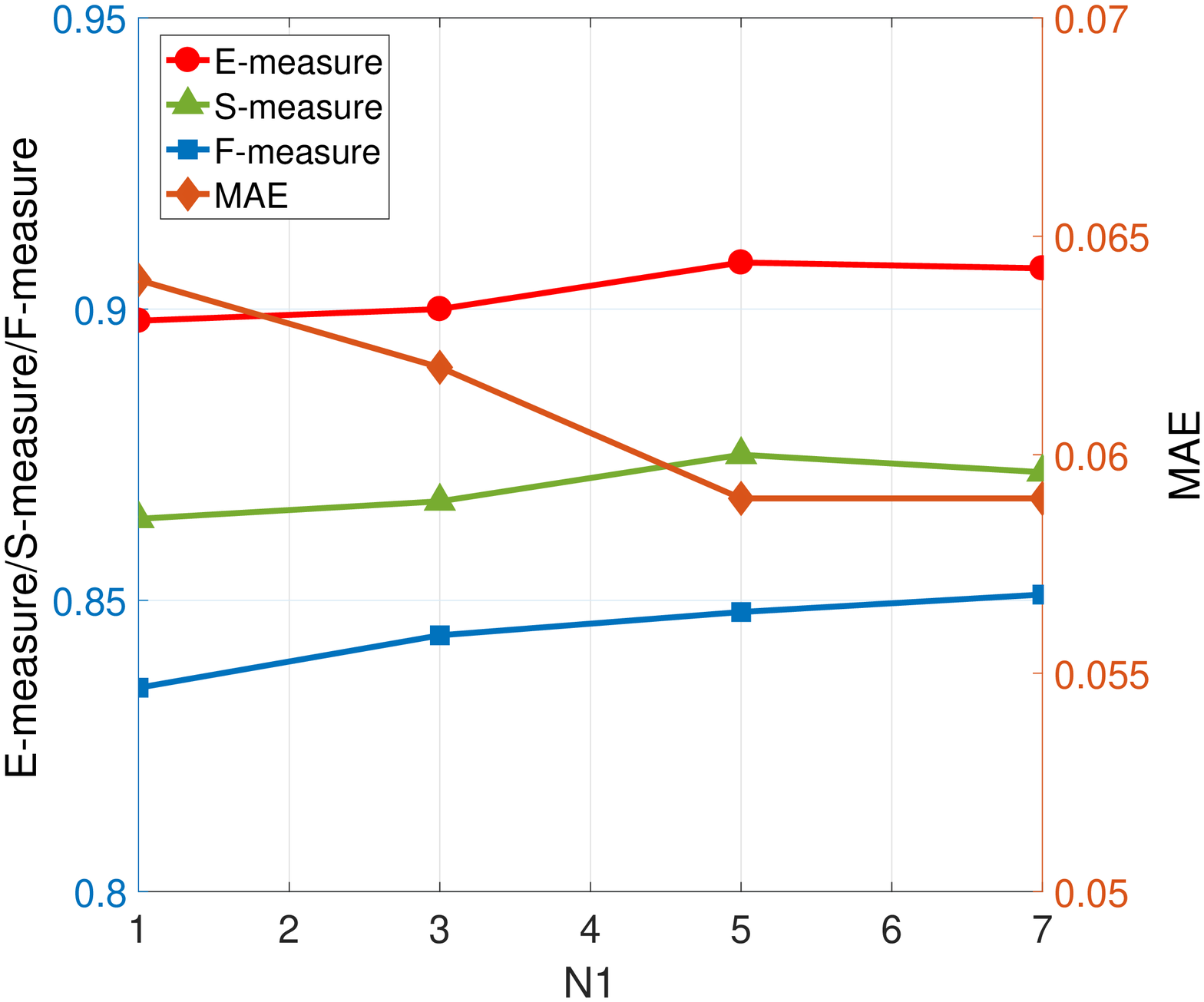}
		\caption{\scriptsize (f) SIP~\cite{fan2019rethinking}}
	\end{subfigure}

    \caption{(a)-(e): Precision-recall curves comparison. (f) Quantitative comparison of different recurrent iterations in the proposed MSR block.}\label{fig_pr}
\end{figure}

\textbf{Quantitative Evaluation.} The quantitative comparison results in terms of 4 evaluation metrics on 7 datasets are reported in Table~\ref{table:qua}. As can be clearly observed that the proposed network significantly outperforms the competing methods across all the datasets in all the metrics except $E_{\xi}$. Comparing with the recent state-of-the-art model DMRA~\cite{piao2019depth}, our approach increases its $S_{\alpha}$ and $F_{\beta}$ scores by an average of \textbf{2.8\%} and \textbf{2.1\%}, decreases the $M$ by an average of \textbf{1.0\%}, which clearly indicates the good consistence with the ground truth. We also perform much better when comparing with CPFP~\cite{zhao2019contrast} which doesn't use pre-trained backbone network to extract depth features too. We analyze that their proposed contrast prior may also break the good property of the RGB features when the depth image is low quality, while such issue can be well alleviated by our method. It is also worth to note that our model trained on NJUD+NLPR still performs better than DMRA on some datasets, which further demonstrates the superiority and effectiveness of the proposed approach. We also plot the PR curves for comparison on five large datasets. As illustrated in Fig.~\ref{fig_pr}, we consistently achieve the best performance especially at a high level of recall.

\setlength{\tabcolsep}{6pt}
\begin{table}[]
\begin{center}
\caption{Running speed and model size comparisons with recent models.}
\label{table_time}
\begin{tabular}{lcccc}
\hline
\hline
Method & Platform & Image Size & FPS$~\uparrow$ & MS (MB)$~\downarrow$ \\ \hline
CPFP~\cite{zhao2019contrast}   & Caffe    & 400$\times$300     &  10   & 291.9       \\
DMRA~\cite{piao2019depth}   & Pytorch  & 256$\times$256     & 16  & 238.8     \\
Ours   & Pytorch  & 352$\times$352     & \textbf{71}  & \textbf{64.9}      \\ \hline
\hline
\end{tabular}
\end{center}
\end{table}
\setlength{\tabcolsep}{1.6pt}

\begin{figure}[t]
\captionsetup[subfigure]{labelformat=empty}	
	\centering
	\begin{subfigure}[t]{1.13cm}
		\centering
		\includegraphics[width=1.13cm]{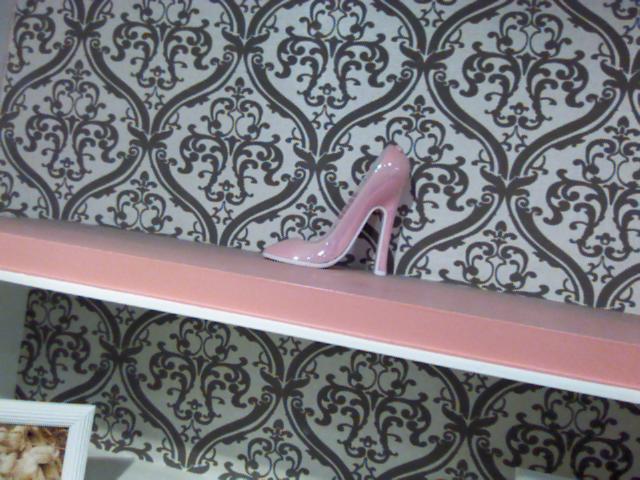}
	\end{subfigure}
	\begin{subfigure}[t]{1.13cm}
		\centering
		\includegraphics[width=1.13cm]{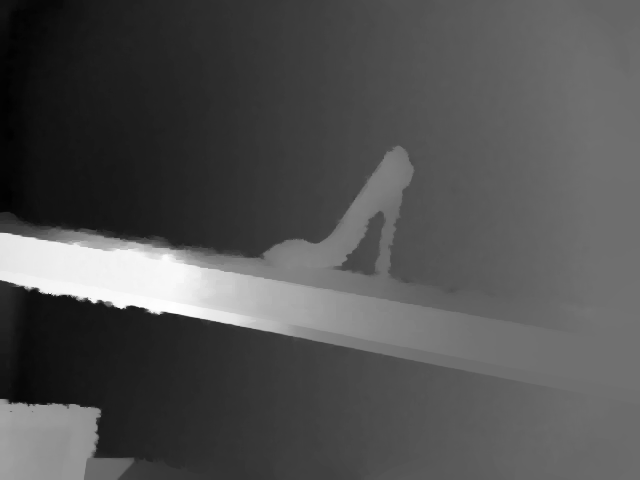}
	\end{subfigure}
	\begin{subfigure}[t]{1.13cm}
		\centering
		\includegraphics[width=1.13cm]{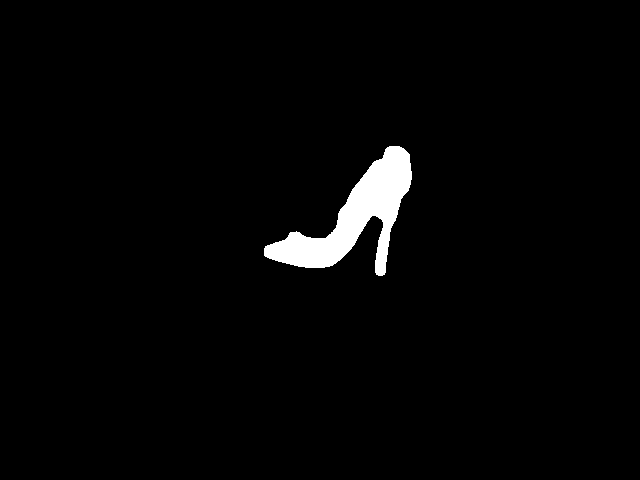}
	\end{subfigure}
	\begin{subfigure}[t]{1.13cm}
		\centering
		\includegraphics[width=1.13cm]{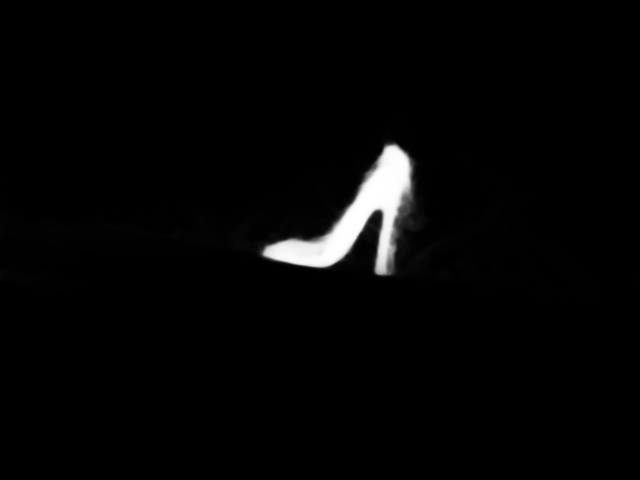}
	\end{subfigure}
	\begin{subfigure}[t]{1.13cm}
		\centering
		\includegraphics[width=1.13cm]{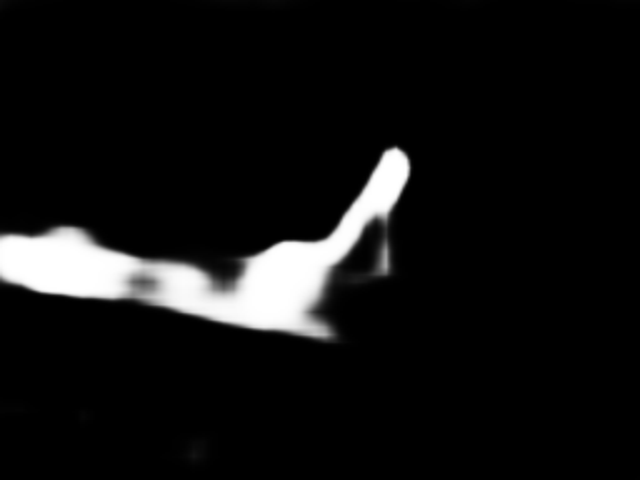}
	\end{subfigure}
	\begin{subfigure}[t]{1.13cm}
		\centering
		\includegraphics[width=1.13cm]{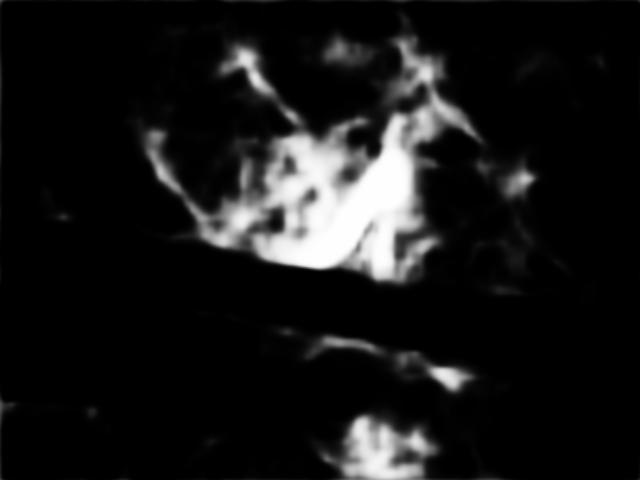}
	\end{subfigure}
	\begin{subfigure}[t]{1.13cm}
		\centering
		\includegraphics[width=1.13cm]{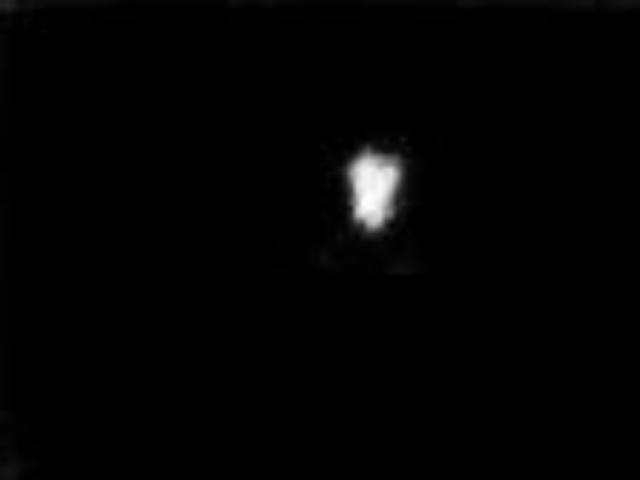}
	\end{subfigure}
	\begin{subfigure}[t]{1.13cm}
		\centering
		\includegraphics[width=1.13cm]{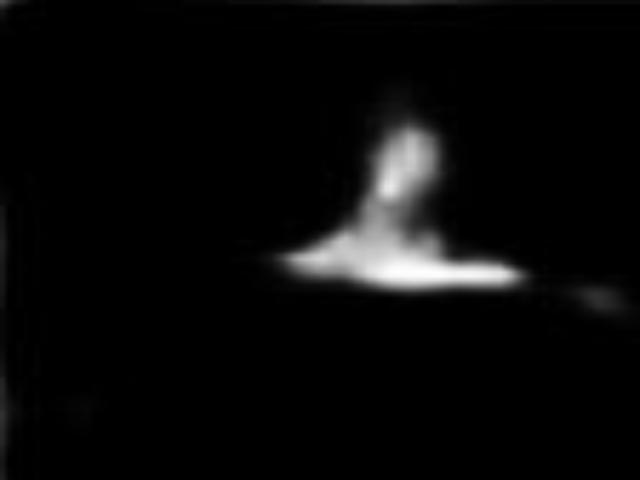}
	\end{subfigure}
	\begin{subfigure}[t]{1.13cm}
		\centering
		\includegraphics[width=1.13cm]{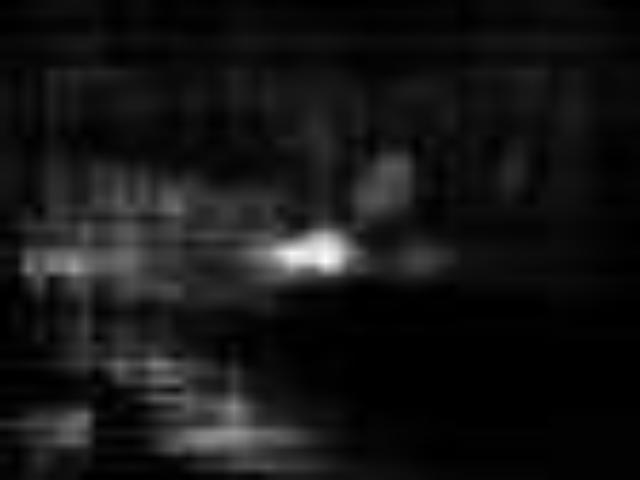}
	\end{subfigure}
	\begin{subfigure}[t]{1.13cm}
		\centering
		\includegraphics[width=1.13cm]{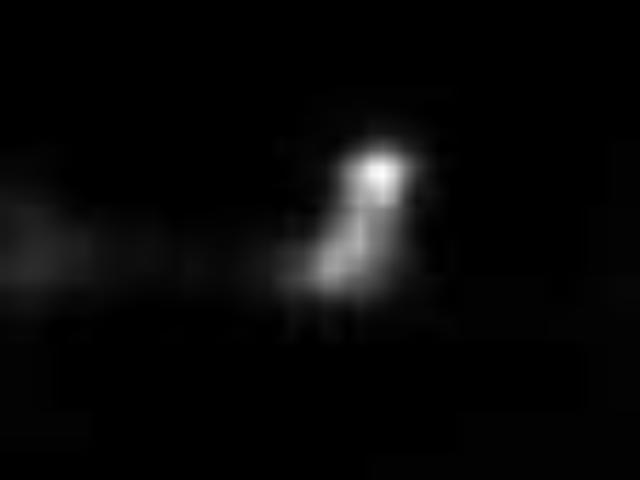}
	\end{subfigure}
	
	\vspace{1pt}
	
	\begin{subfigure}[t]{1.13cm}
		\centering
		\includegraphics[width=1.13cm]{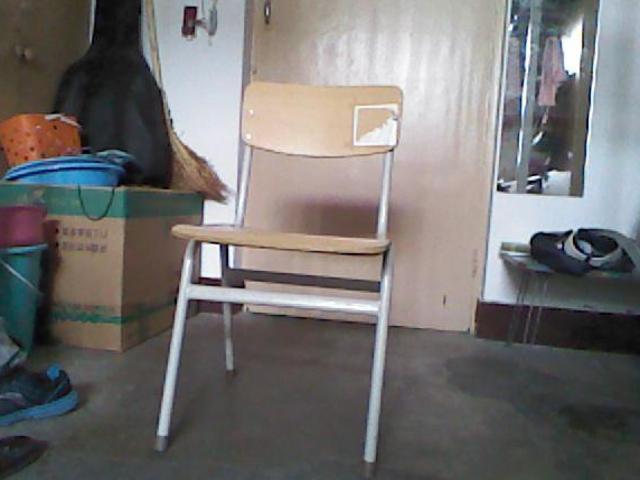}
	\end{subfigure}
	\begin{subfigure}[t]{1.13cm}
		\centering
		\includegraphics[width=1.13cm]{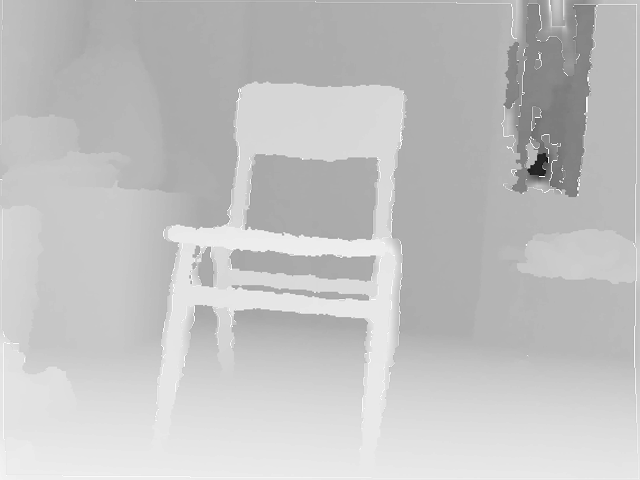}
	\end{subfigure}
	\begin{subfigure}[t]{1.13cm}
		\centering
		\includegraphics[width=1.13cm]{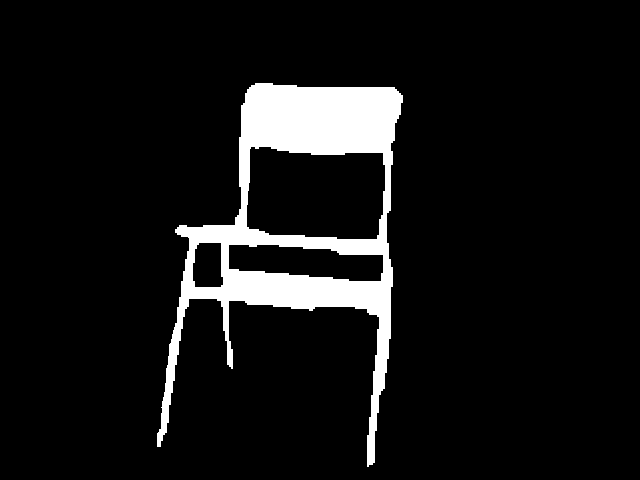}
	\end{subfigure}
	\begin{subfigure}[t]{1.13cm}
		\centering
		\includegraphics[width=1.13cm]{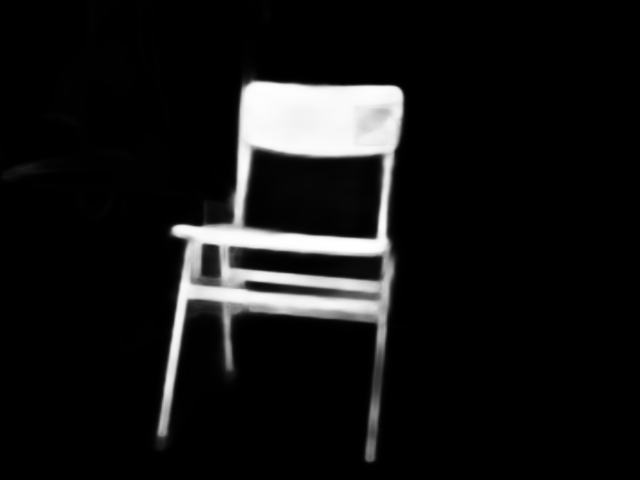}
	\end{subfigure}
	\begin{subfigure}[t]{1.13cm}
		\centering
		\includegraphics[width=1.13cm]{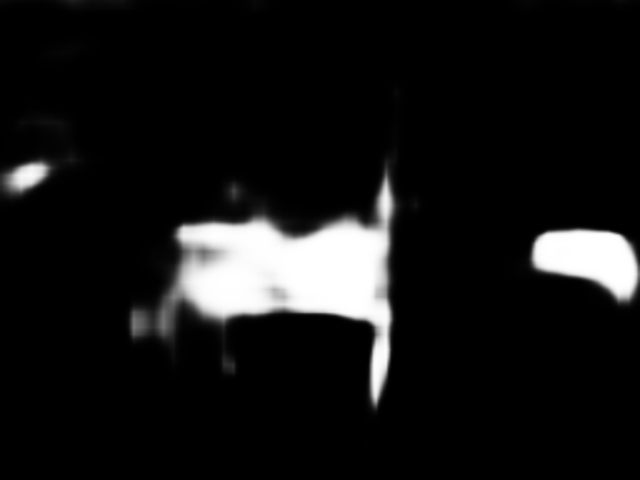}
	\end{subfigure}
	\begin{subfigure}[t]{1.13cm}
		\centering
		\includegraphics[width=1.13cm]{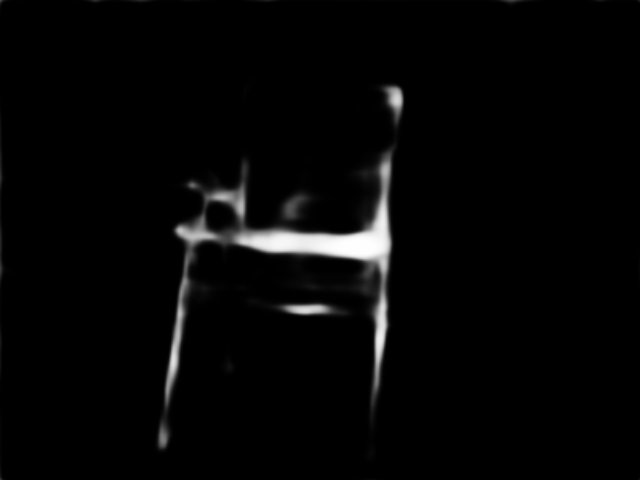}
	\end{subfigure}
	\begin{subfigure}[t]{1.13cm}
		\centering
		\includegraphics[width=1.13cm]{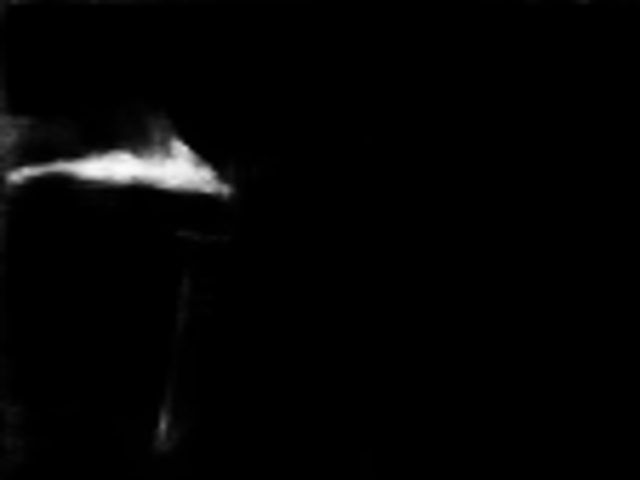}
	\end{subfigure}
	\begin{subfigure}[t]{1.13cm}
		\centering
		\includegraphics[width=1.13cm]{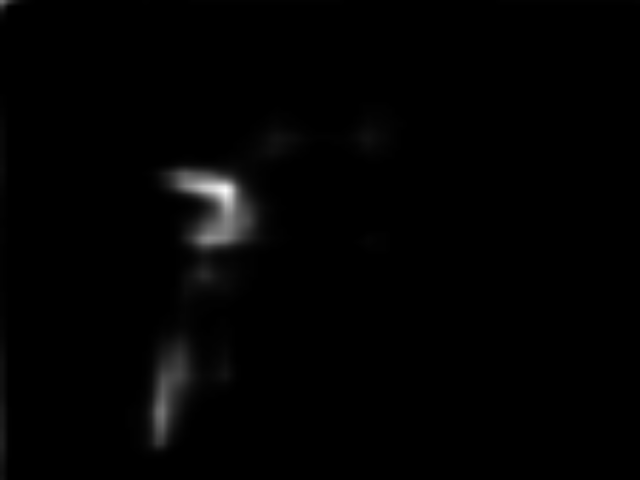}
	\end{subfigure}
	\begin{subfigure}[t]{1.13cm}
		\centering
		\includegraphics[width=1.13cm]{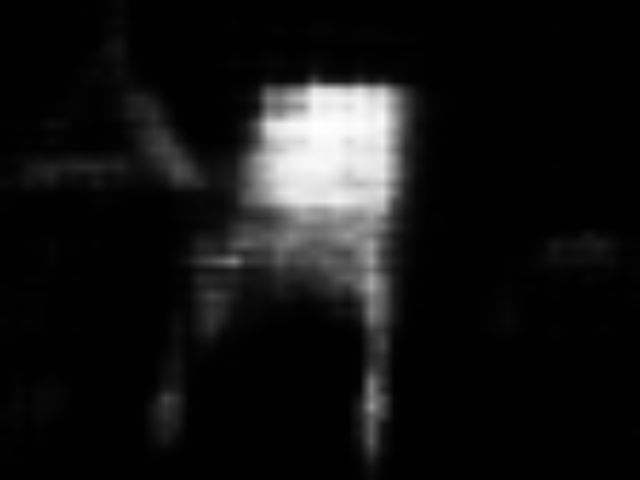}
	\end{subfigure}
	\begin{subfigure}[t]{1.13cm}
		\centering
		\includegraphics[width=1.13cm]{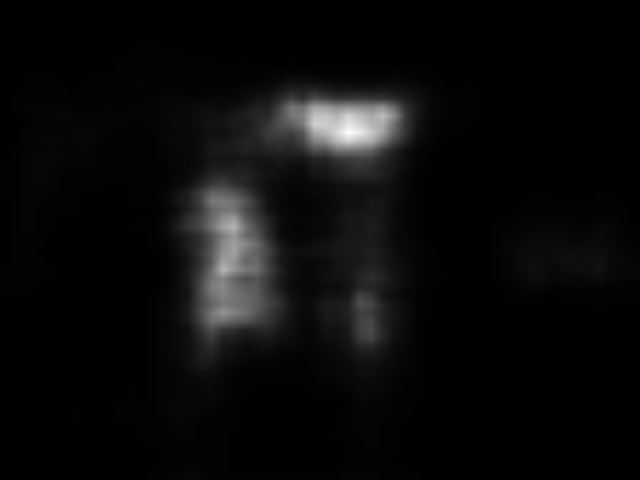}
	\end{subfigure}
	
	\vspace{1pt}
	
	\begin{subfigure}[t]{1.13cm}
		\centering
		\includegraphics[width=1.13cm]{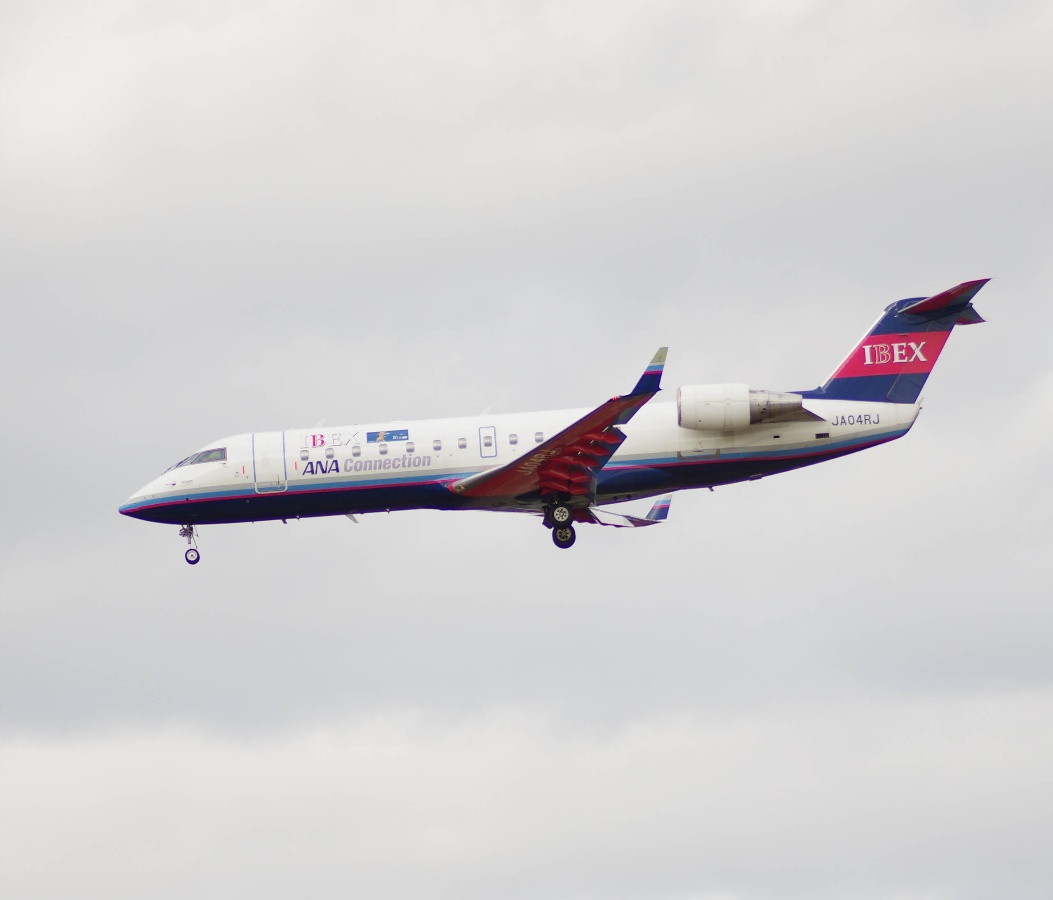}
	\end{subfigure}
	\begin{subfigure}[t]{1.13cm}
		\centering
		\includegraphics[width=1.13cm]{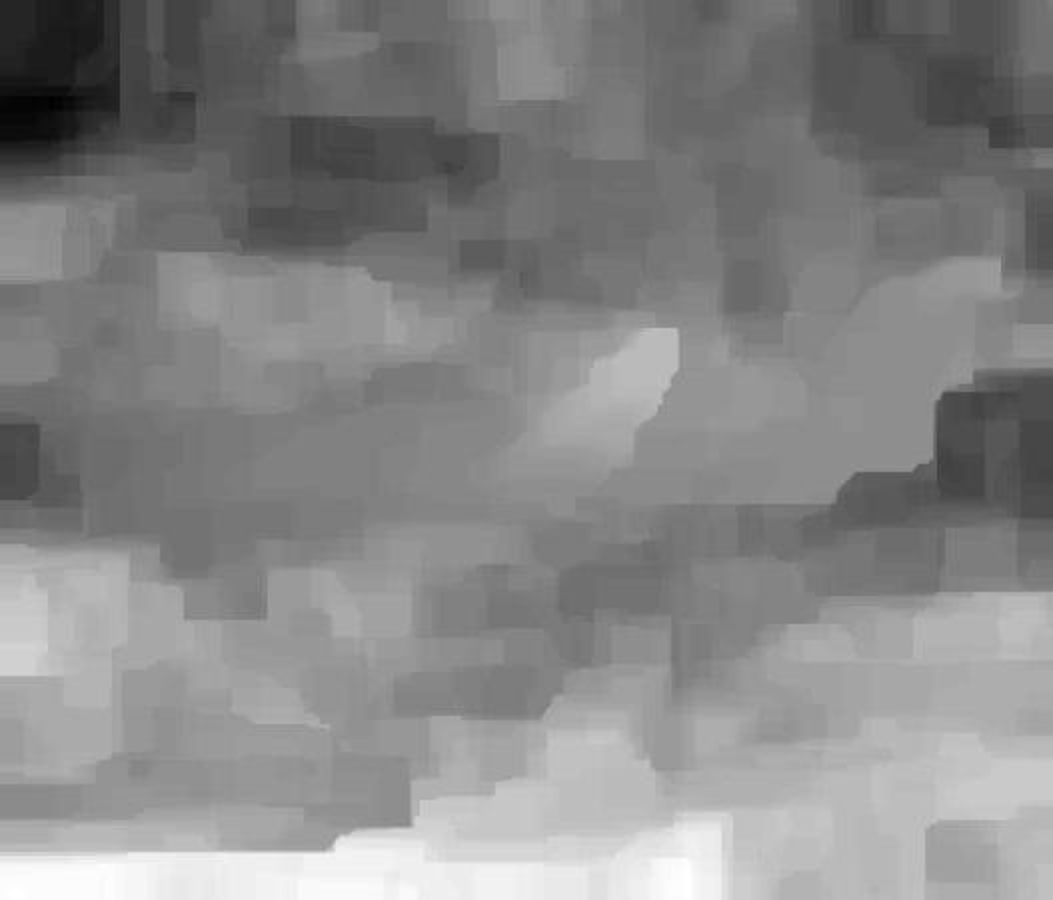}
	\end{subfigure}
	\begin{subfigure}[t]{1.13cm}
		\centering
		\includegraphics[width=1.13cm]{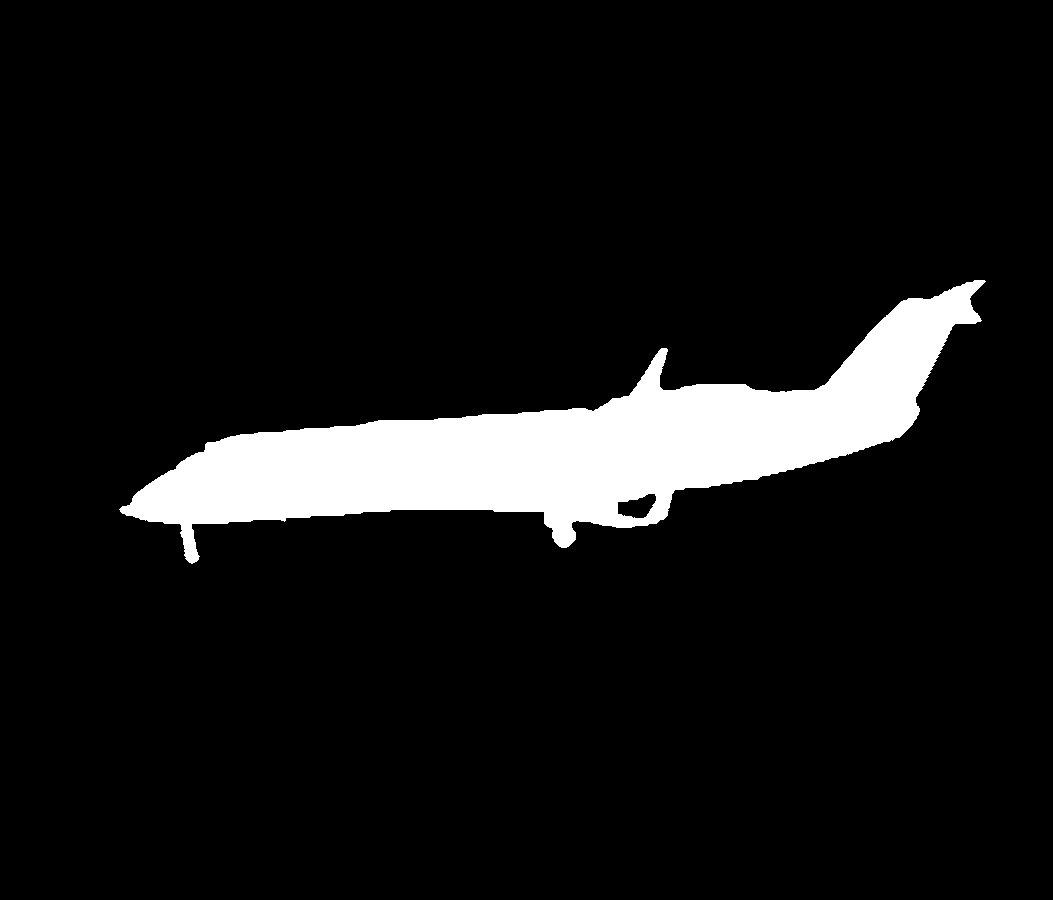}
	\end{subfigure}
	\begin{subfigure}[t]{1.13cm}
		\centering
		\includegraphics[width=1.13cm]{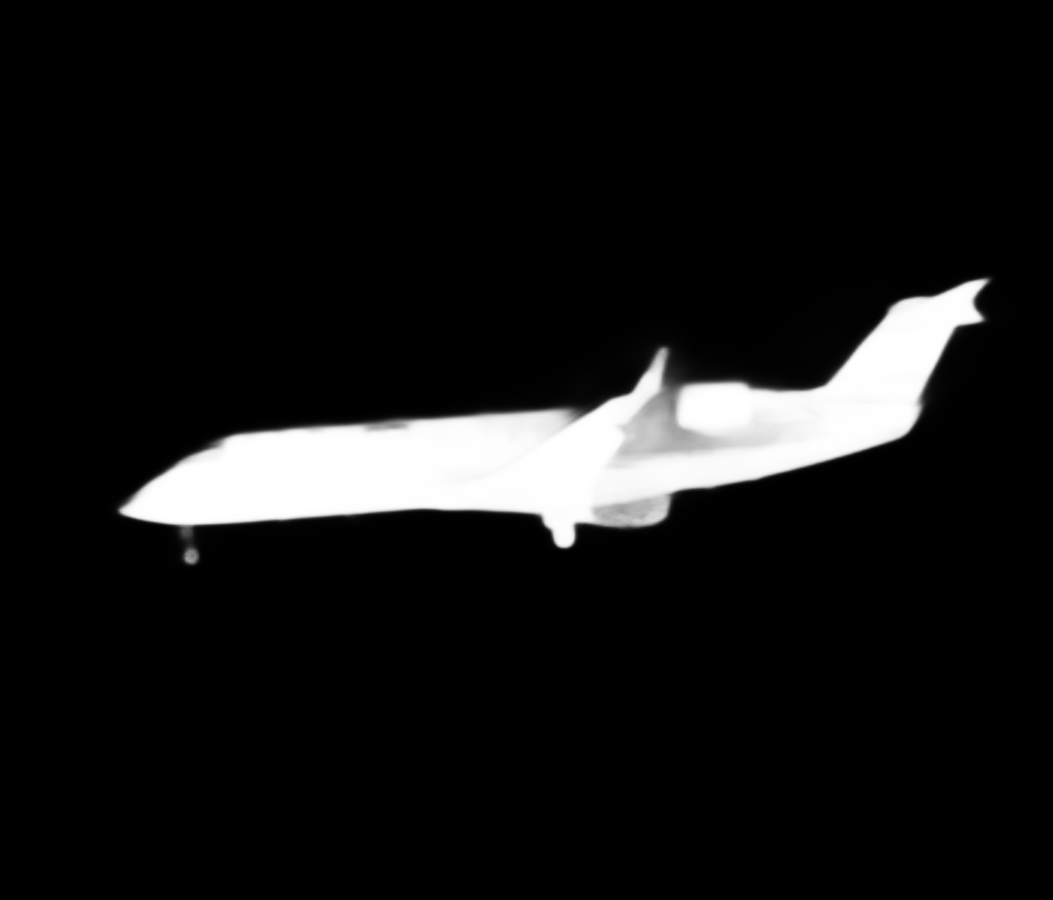}
	\end{subfigure}
	\begin{subfigure}[t]{1.13cm}
		\centering
		\includegraphics[width=1.13cm]{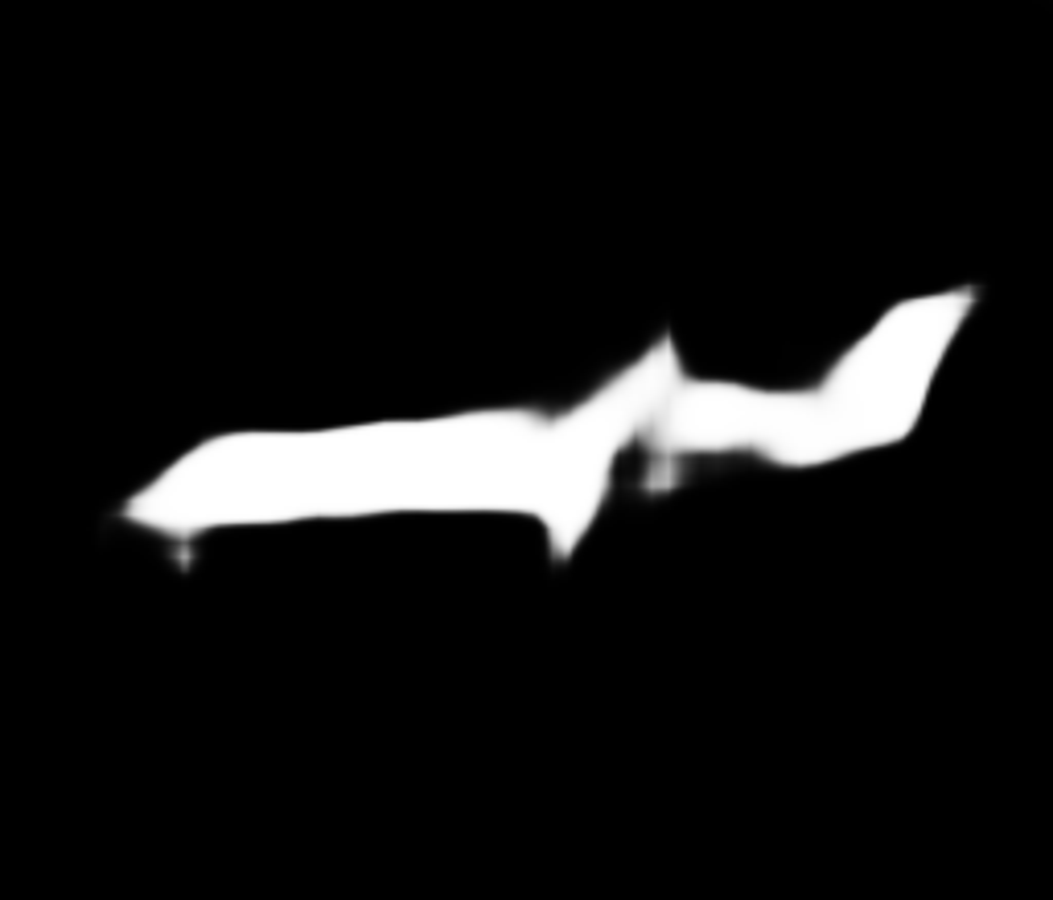}
	\end{subfigure}
	\begin{subfigure}[t]{1.13cm}
		\centering
		\includegraphics[width=1.13cm]{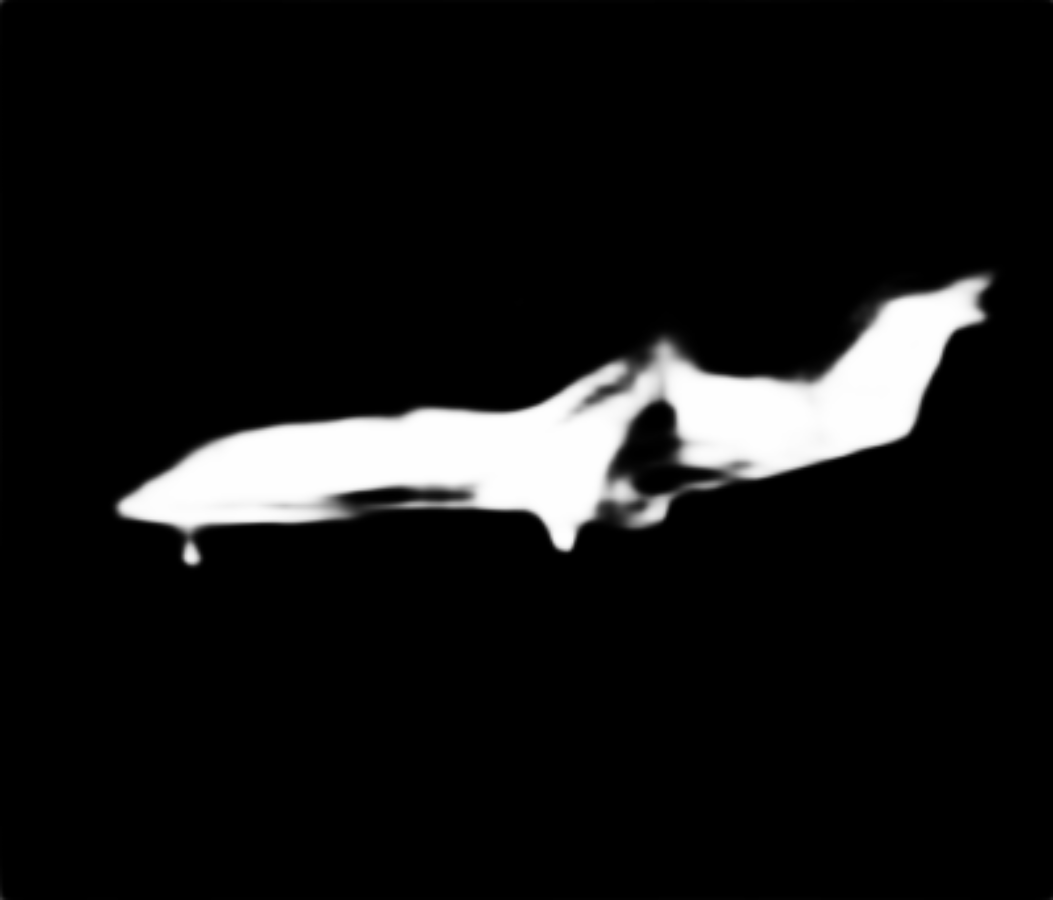}
	\end{subfigure}
	\begin{subfigure}[t]{1.13cm}
		\centering
		\includegraphics[width=1.13cm]{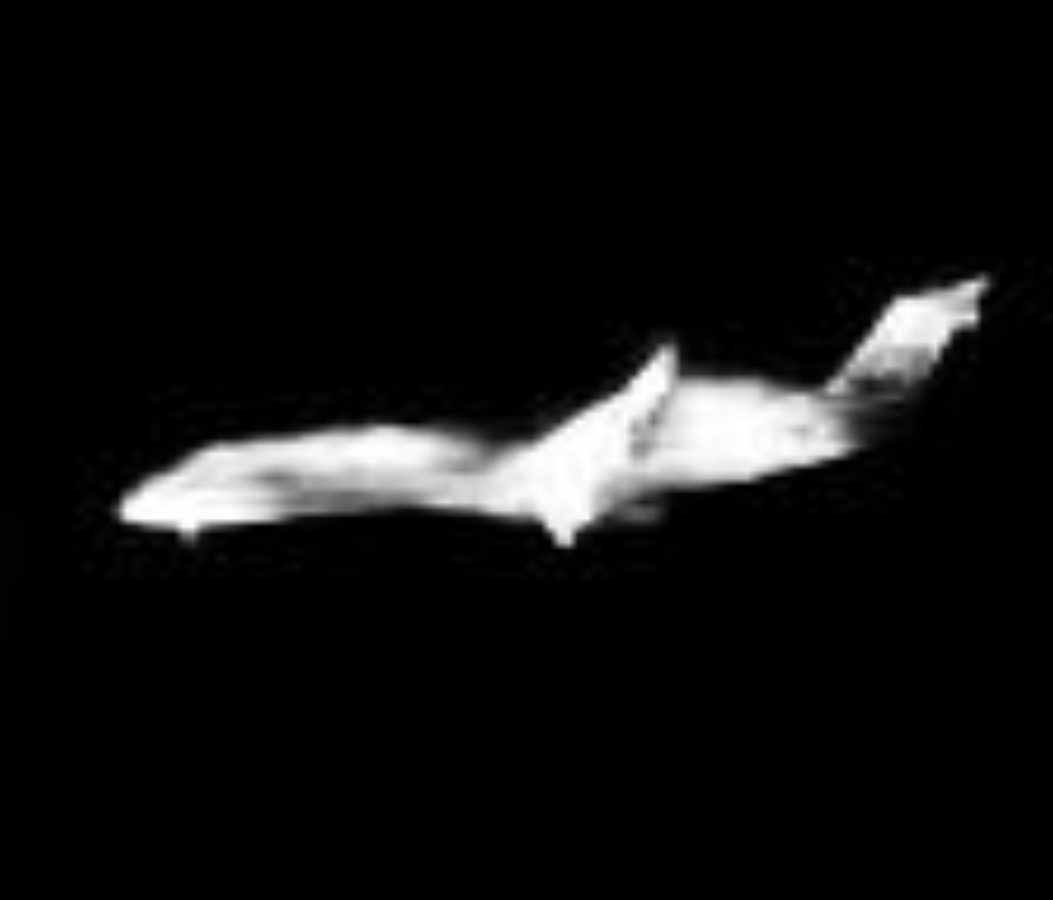}
	\end{subfigure}
	\begin{subfigure}[t]{1.13cm}
		\centering
		\includegraphics[width=1.13cm]{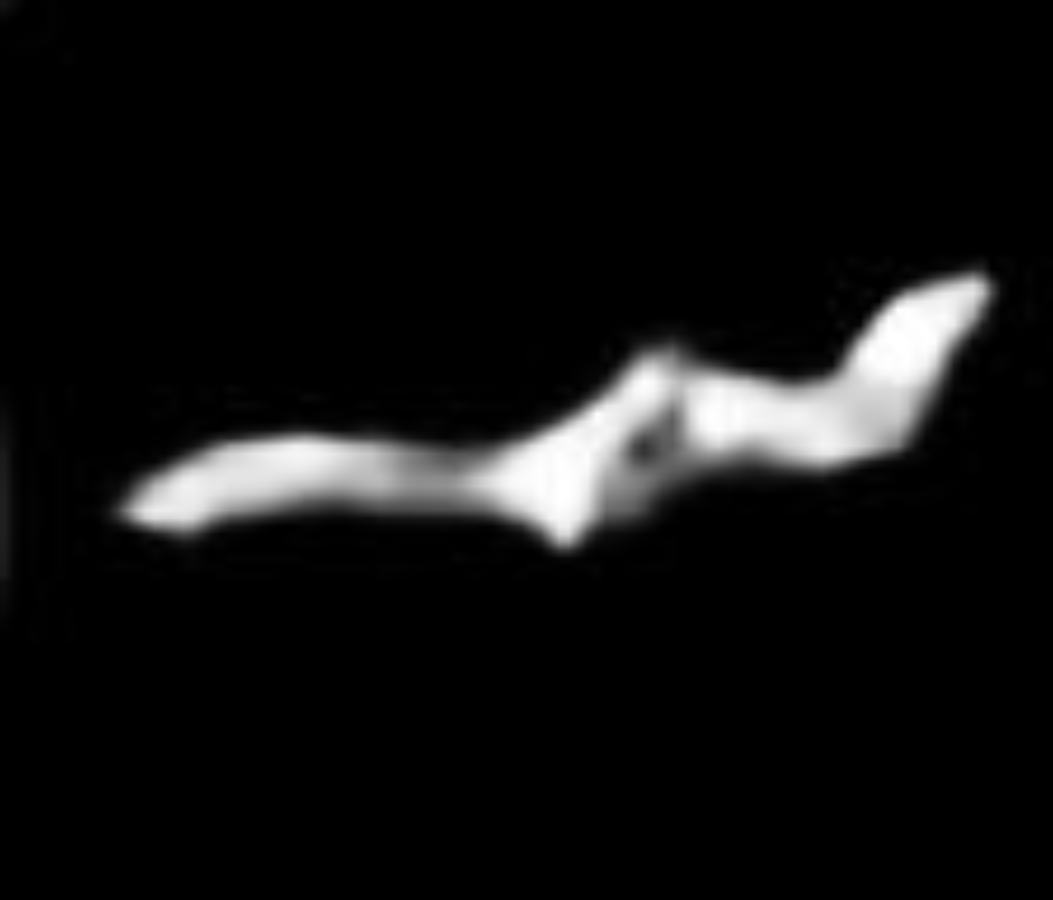}
	\end{subfigure}
	\begin{subfigure}[t]{1.13cm}
		\centering
		\includegraphics[width=1.13cm]{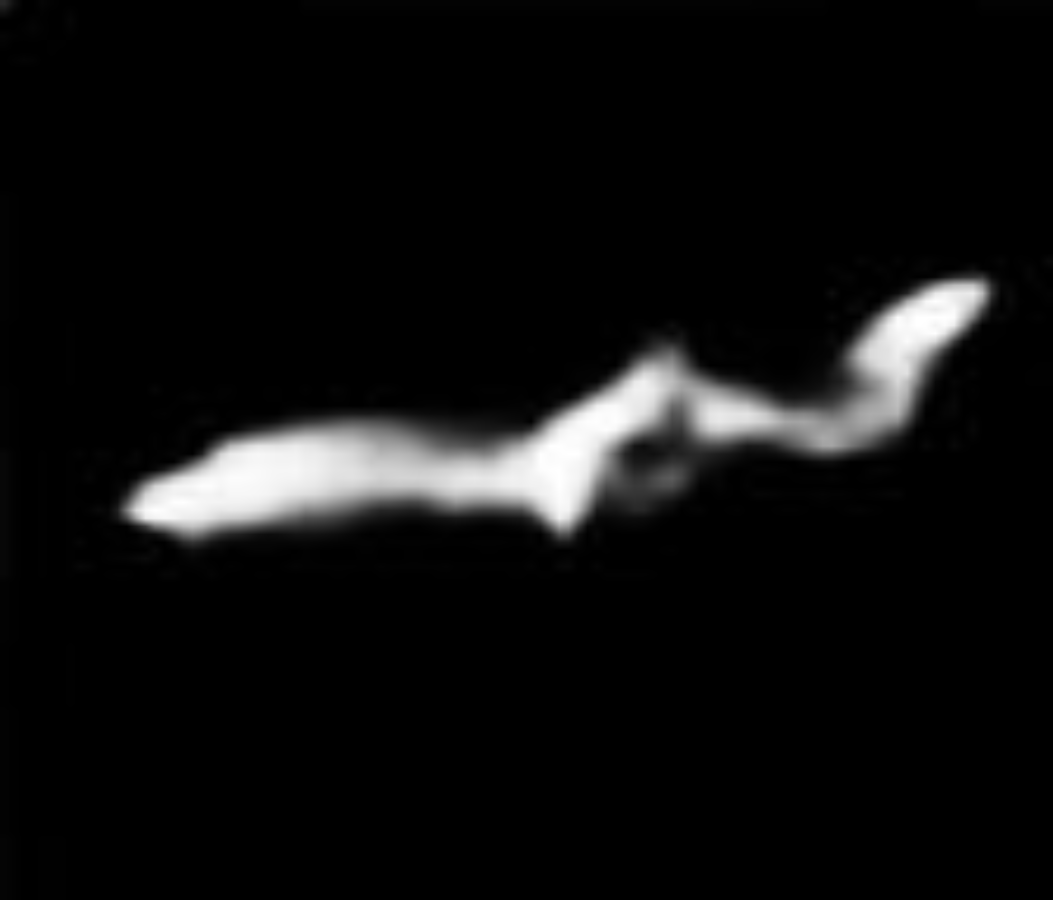}
	\end{subfigure}
	\begin{subfigure}[t]{1.13cm}
		\centering
		\includegraphics[width=1.13cm]{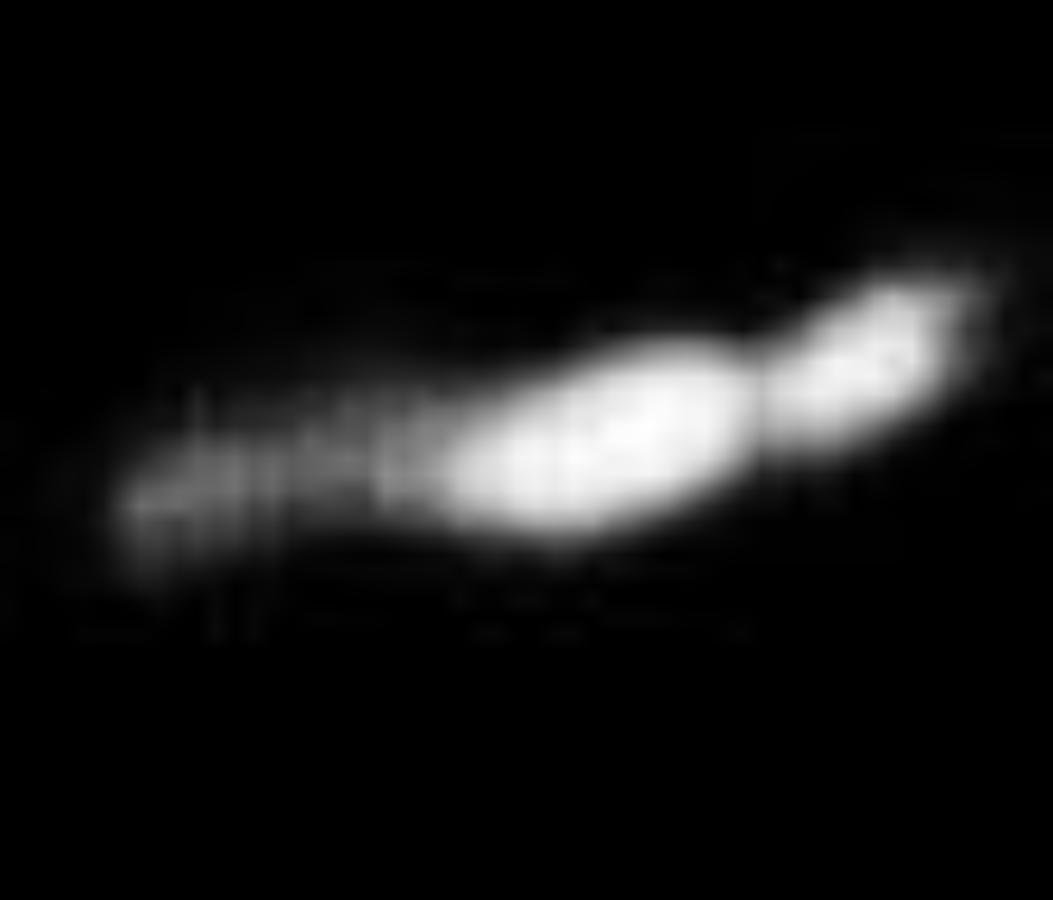}
	\end{subfigure}
	
	\vspace{1pt}
	
	\begin{subfigure}[t]{1.13cm}
		\centering
		\includegraphics[width=1.13cm]{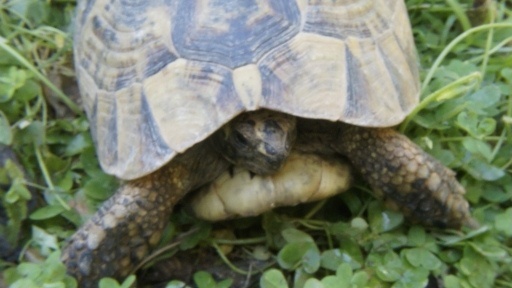}
	\end{subfigure}
	\begin{subfigure}[t]{1.13cm}
		\centering
		\includegraphics[width=1.13cm]{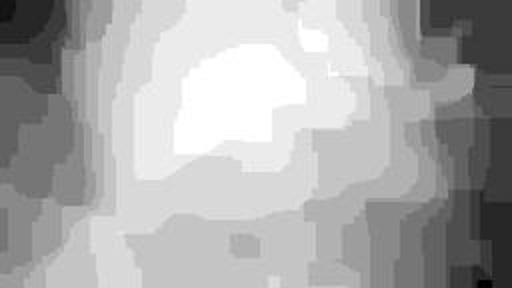}
	\end{subfigure}
	\begin{subfigure}[t]{1.13cm}
		\centering
		\includegraphics[width=1.13cm]{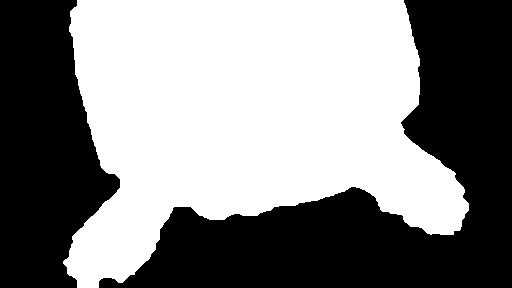}
	\end{subfigure}
	\begin{subfigure}[t]{1.13cm}
		\centering
		\includegraphics[width=1.13cm]{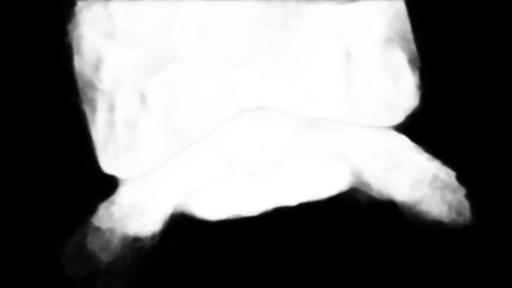}
	\end{subfigure}
	\begin{subfigure}[t]{1.13cm}
		\centering
		\includegraphics[width=1.13cm]{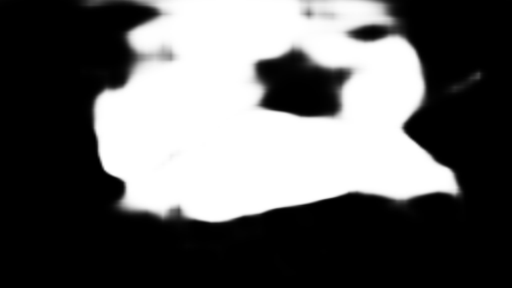}
	\end{subfigure}
	\begin{subfigure}[t]{1.13cm}
		\centering
		\includegraphics[width=1.13cm]{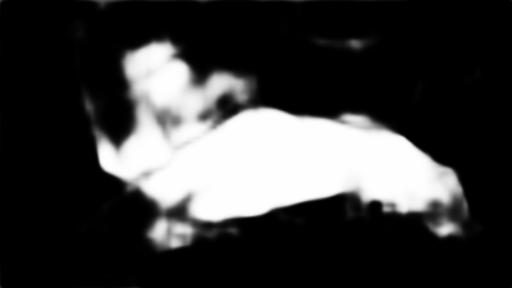}
	\end{subfigure}
	\begin{subfigure}[t]{1.13cm}
		\centering
		\includegraphics[width=1.13cm]{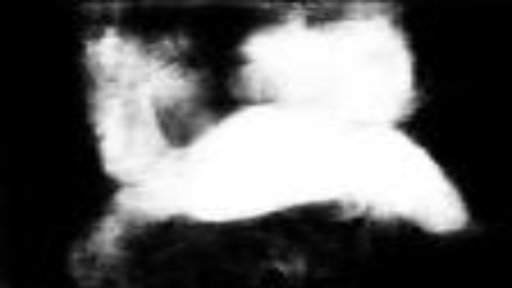}
	\end{subfigure}
	\begin{subfigure}[t]{1.13cm}
		\centering
		\includegraphics[width=1.13cm]{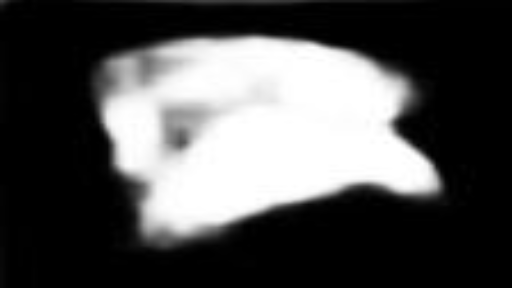}
	\end{subfigure}
	\begin{subfigure}[t]{1.13cm}
		\centering
		\includegraphics[width=1.13cm]{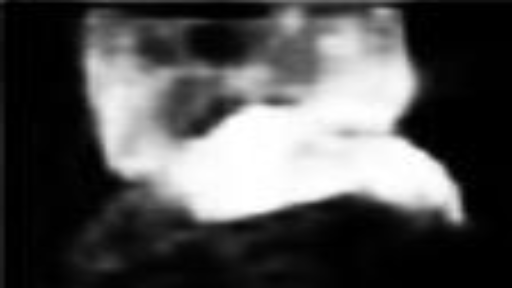}
	\end{subfigure}
	\begin{subfigure}[t]{1.13cm}
		\centering
		\includegraphics[width=1.13cm]{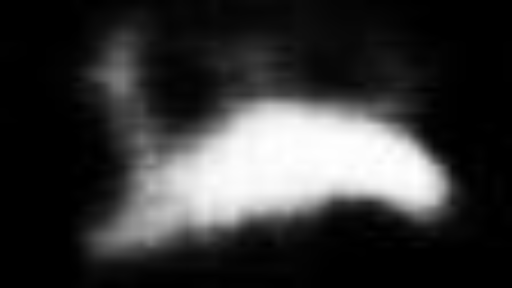}
	\end{subfigure}
	
	\vspace{1pt}
	
	\begin{subfigure}[t]{1.13cm}
		\centering
		\includegraphics[width=1.13cm]{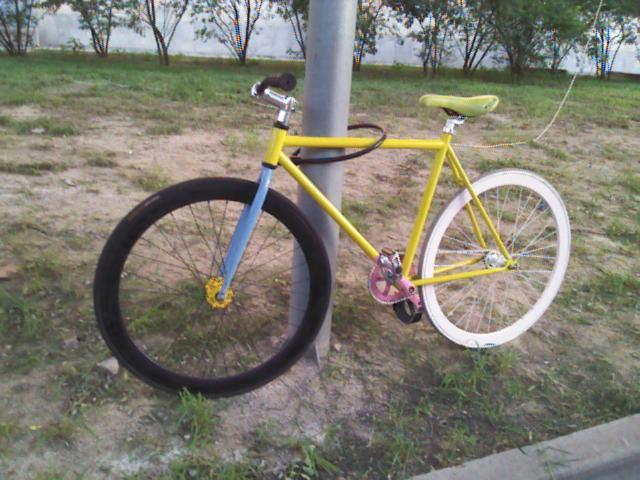}
	\end{subfigure}
	\begin{subfigure}[t]{1.13cm}
		\centering
		\includegraphics[width=1.13cm]{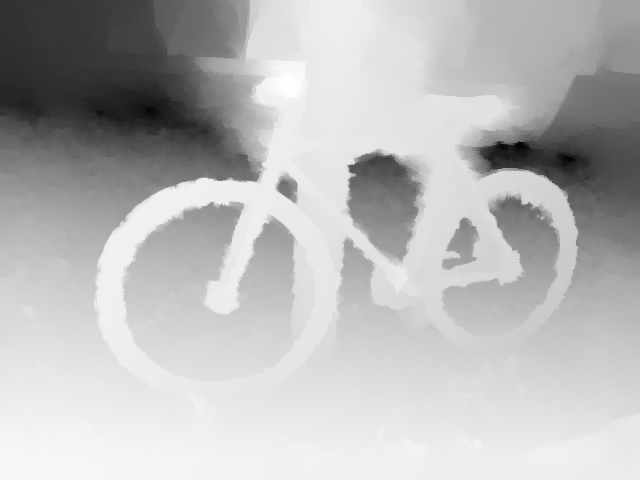}
	\end{subfigure}
	\begin{subfigure}[t]{1.13cm}
		\centering
		\includegraphics[width=1.13cm]{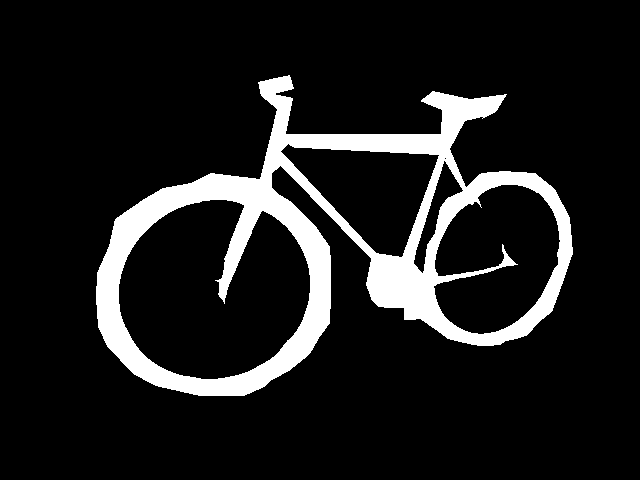}
	\end{subfigure}
	\begin{subfigure}[t]{1.13cm}
		\centering
		\includegraphics[width=1.13cm]{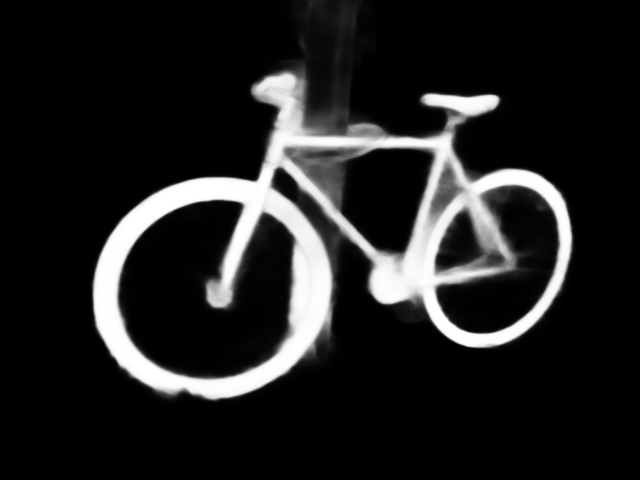}
	\end{subfigure}
	\begin{subfigure}[t]{1.13cm}
		\centering
		\includegraphics[width=1.13cm]{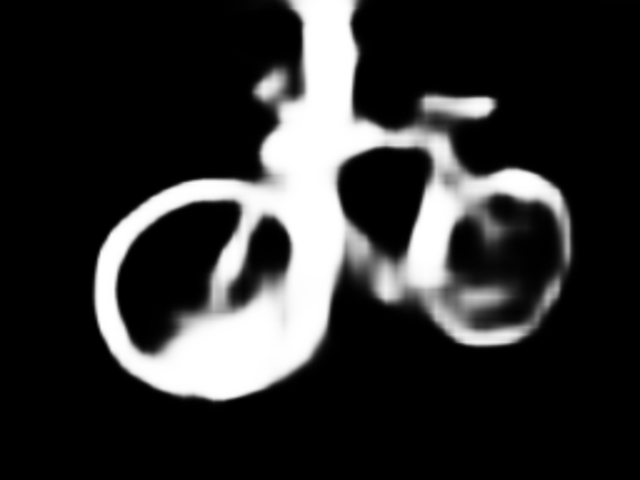}
	\end{subfigure}
	\begin{subfigure}[t]{1.13cm}
		\centering
		\includegraphics[width=1.13cm]{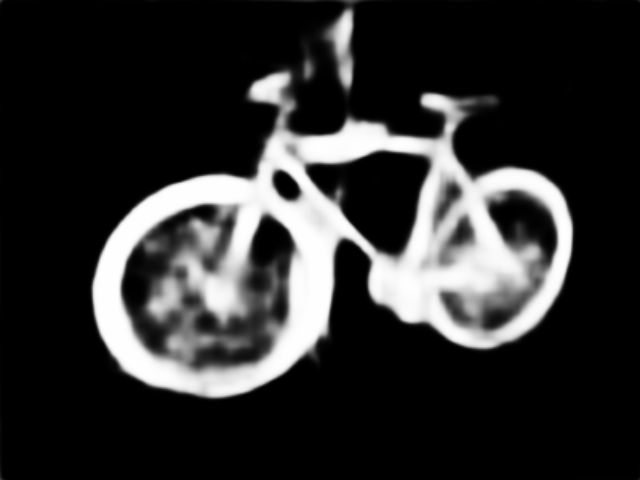}
	\end{subfigure}
	\begin{subfigure}[t]{1.13cm}
		\centering
		\includegraphics[width=1.13cm]{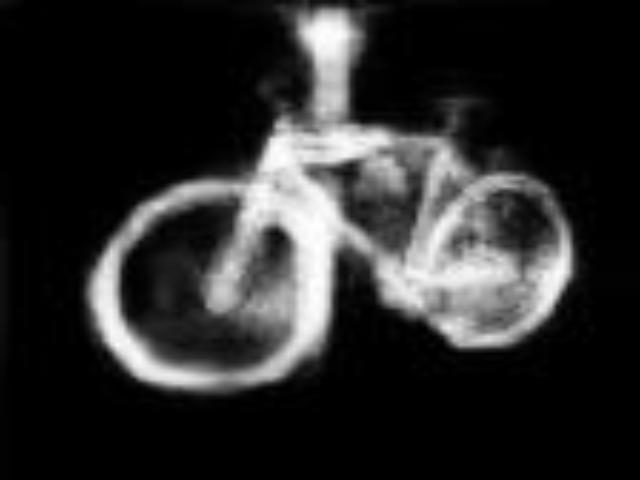}
	\end{subfigure}
	\begin{subfigure}[t]{1.13cm}
		\centering
		\includegraphics[width=1.13cm]{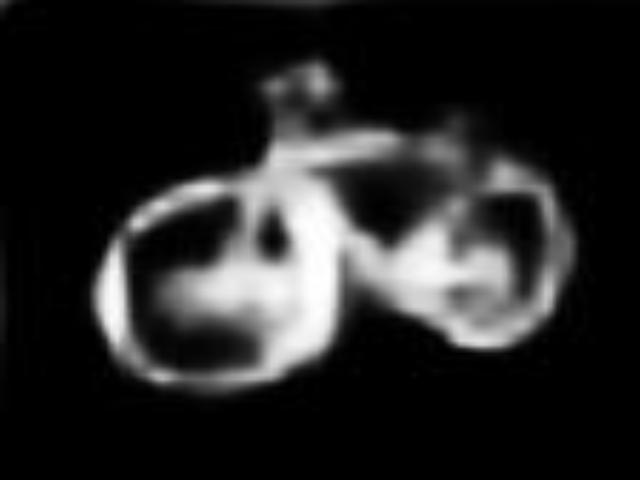}
	\end{subfigure}
	\begin{subfigure}[t]{1.13cm}
		\centering
		\includegraphics[width=1.13cm]{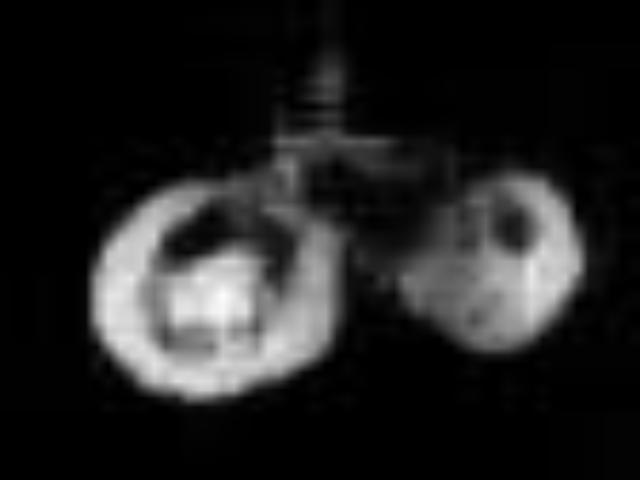}
	\end{subfigure}
	\begin{subfigure}[t]{1.13cm}
		\centering
		\includegraphics[width=1.13cm]{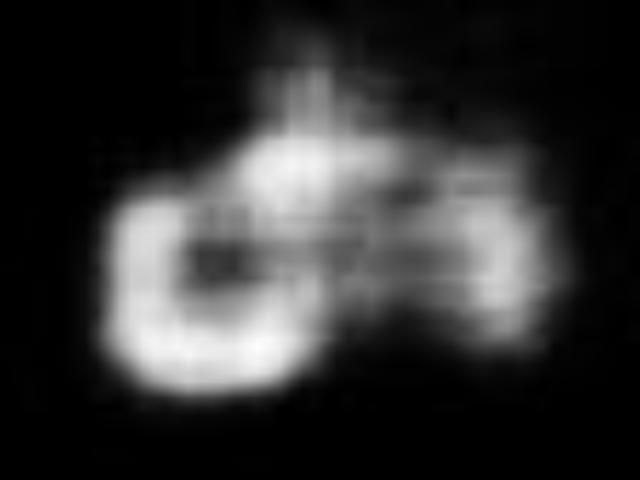}
	\end{subfigure}
	
	\vspace{1pt}
	
	\begin{subfigure}[t]{1.13cm}
		\centering
		\includegraphics[width=1.13cm]{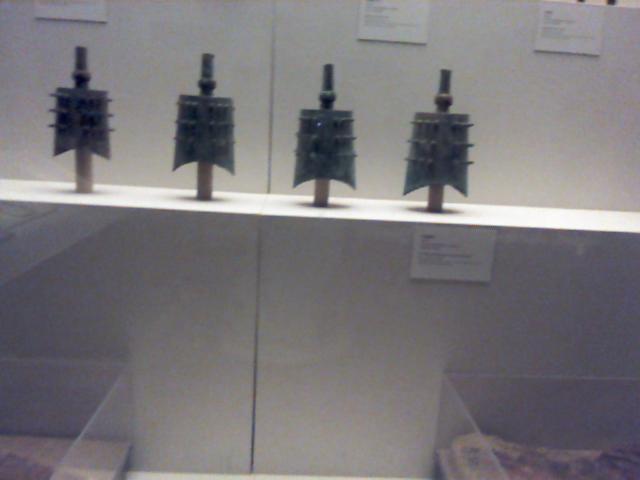}
	\end{subfigure}
	\begin{subfigure}[t]{1.13cm}
		\centering
		\includegraphics[width=1.13cm]{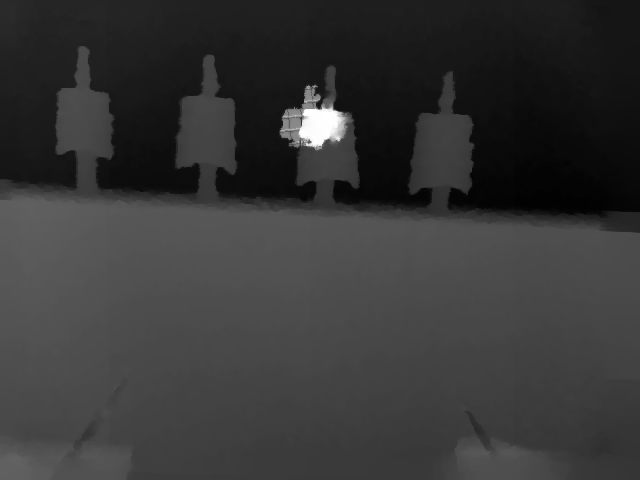}
	\end{subfigure}
	\begin{subfigure}[t]{1.13cm}
		\centering
		\includegraphics[width=1.13cm]{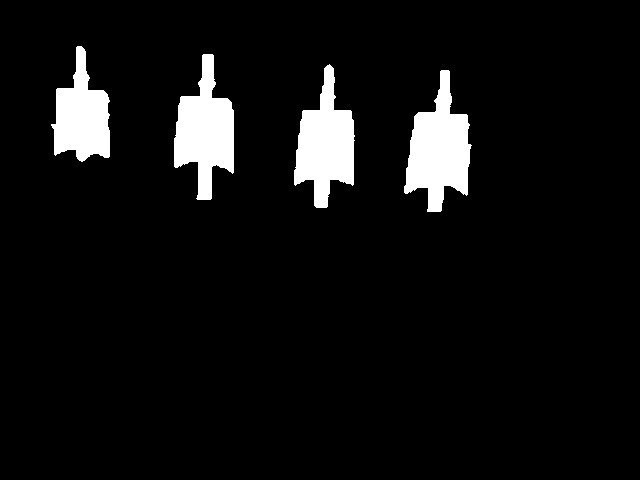}
	\end{subfigure}
	\begin{subfigure}[t]{1.13cm}
		\centering
		\includegraphics[width=1.13cm]{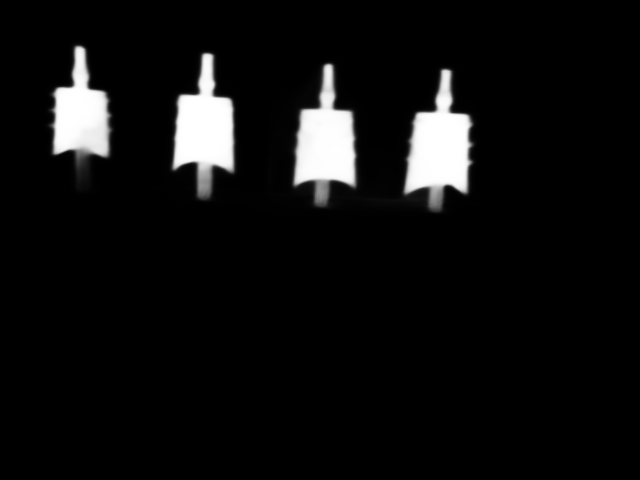}
	\end{subfigure}
	\begin{subfigure}[t]{1.13cm}
		\centering
		\includegraphics[width=1.13cm]{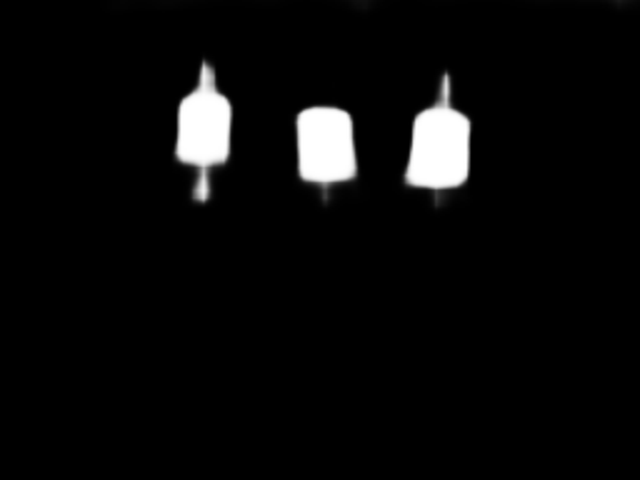}
	\end{subfigure}
	\begin{subfigure}[t]{1.13cm}
		\centering
		\includegraphics[width=1.13cm]{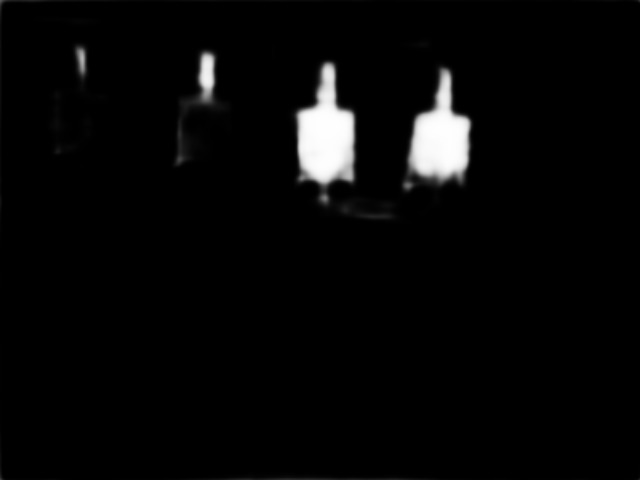}
	\end{subfigure}
	\begin{subfigure}[t]{1.13cm}
		\centering
		\includegraphics[width=1.13cm]{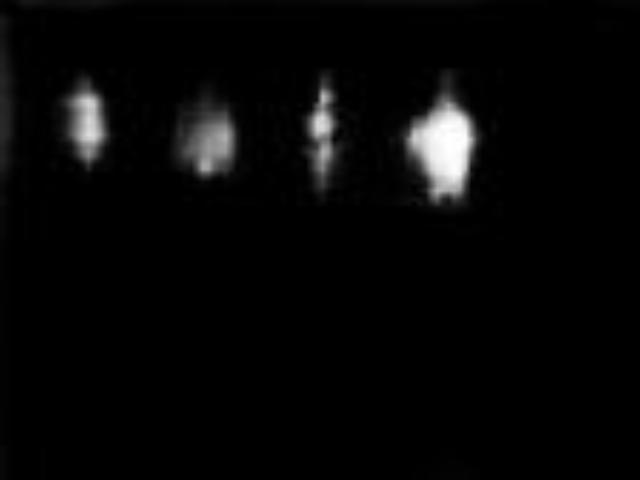}
	\end{subfigure}
	\begin{subfigure}[t]{1.13cm}
		\centering
		\includegraphics[width=1.13cm]{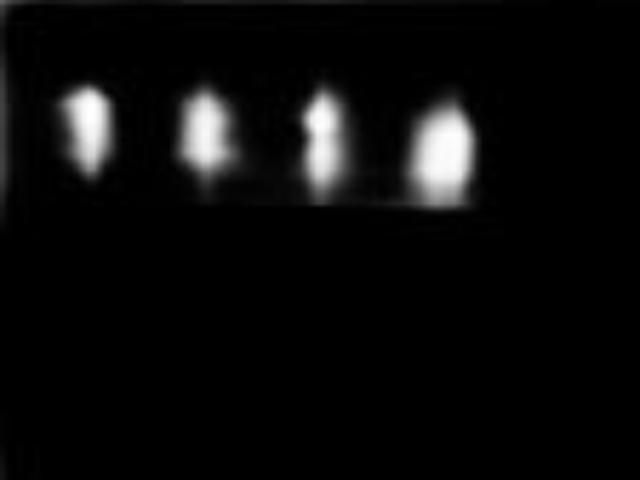}
	\end{subfigure}
	\begin{subfigure}[t]{1.13cm}
		\centering
		\includegraphics[width=1.13cm]{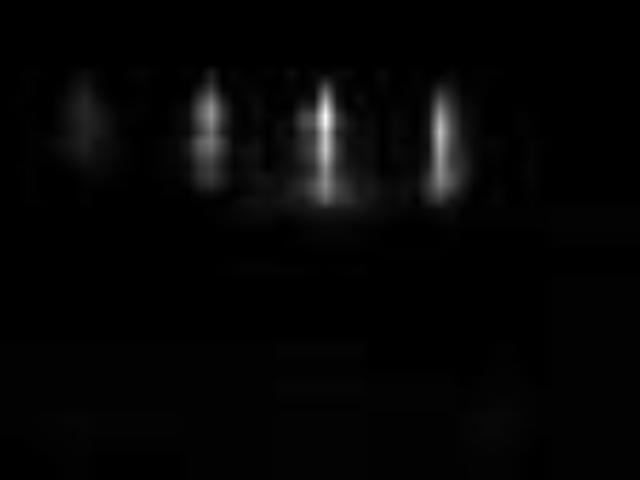}
	\end{subfigure}
	\begin{subfigure}[t]{1.13cm}
		\centering
		\includegraphics[width=1.13cm]{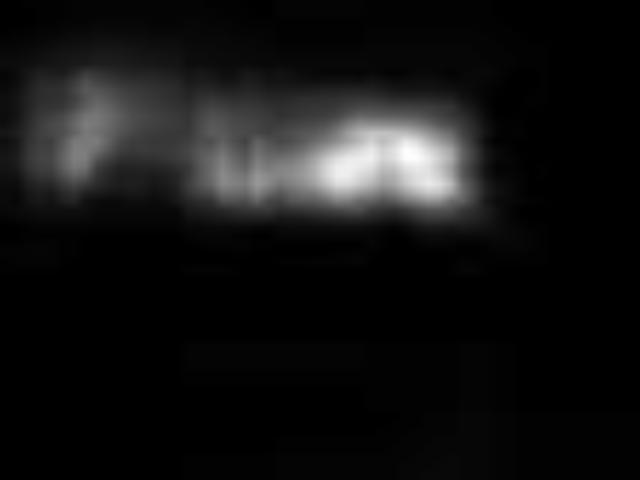}
	\end{subfigure}
	
	\vspace{1pt}
	
	\begin{subfigure}[t]{1.13cm}
		\centering
		\includegraphics[width=1.13cm]{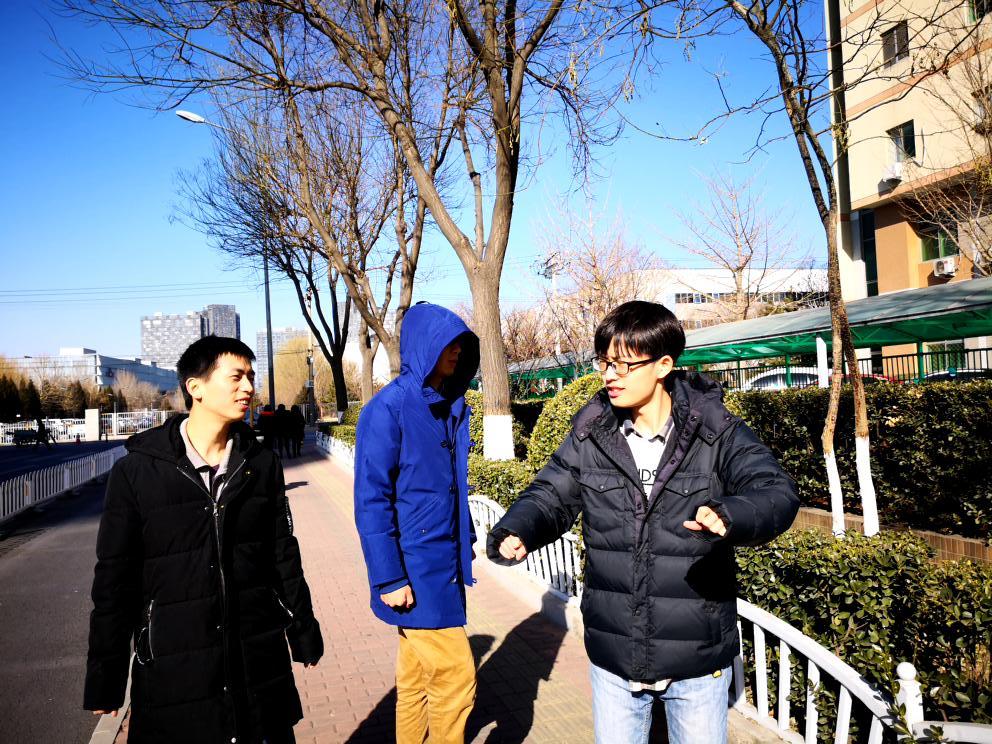}
	\end{subfigure}
	\begin{subfigure}[t]{1.13cm}
		\centering
		\includegraphics[width=1.13cm]{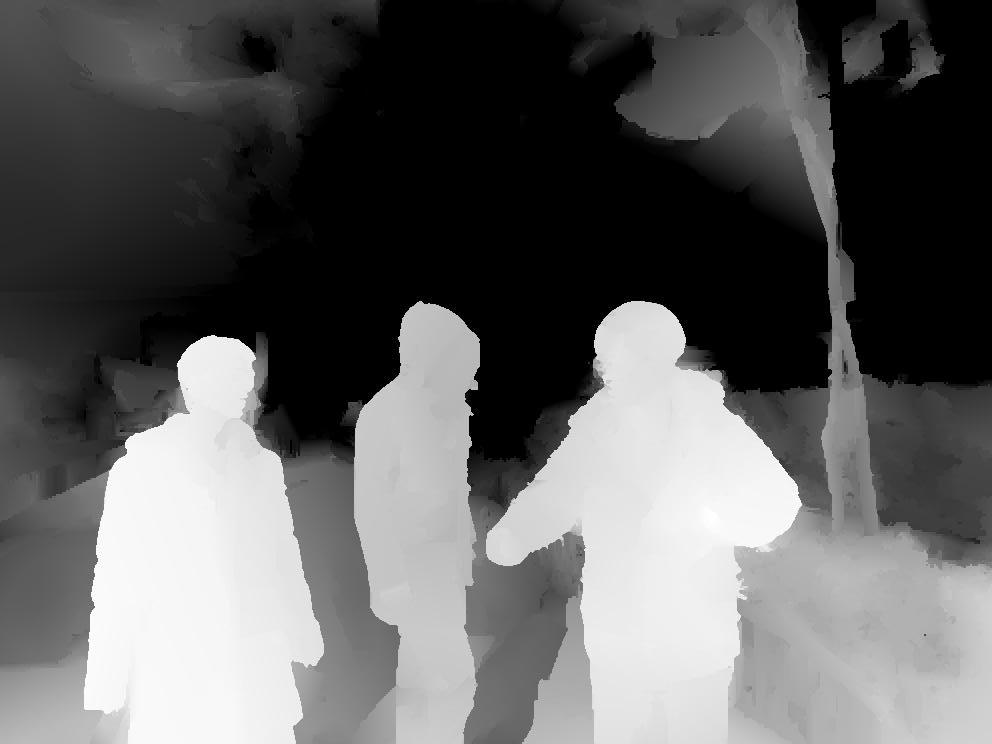}
	\end{subfigure}
	\begin{subfigure}[t]{1.13cm}
		\centering
		\includegraphics[width=1.13cm]{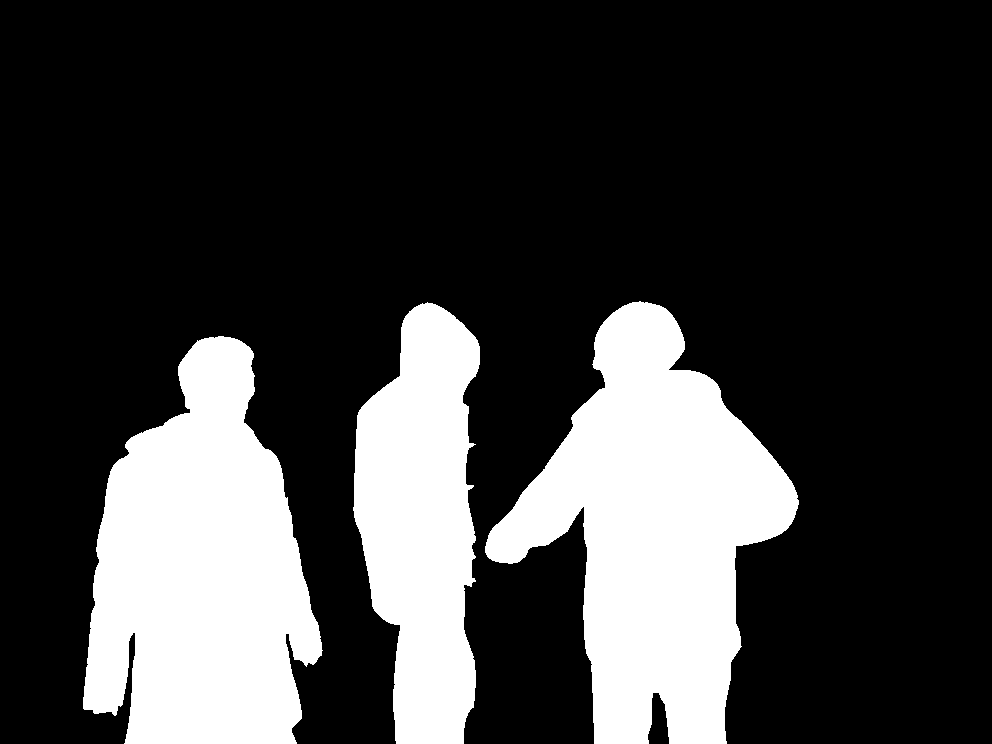}
	\end{subfigure}
	\begin{subfigure}[t]{1.13cm}
		\centering
		\includegraphics[width=1.13cm]{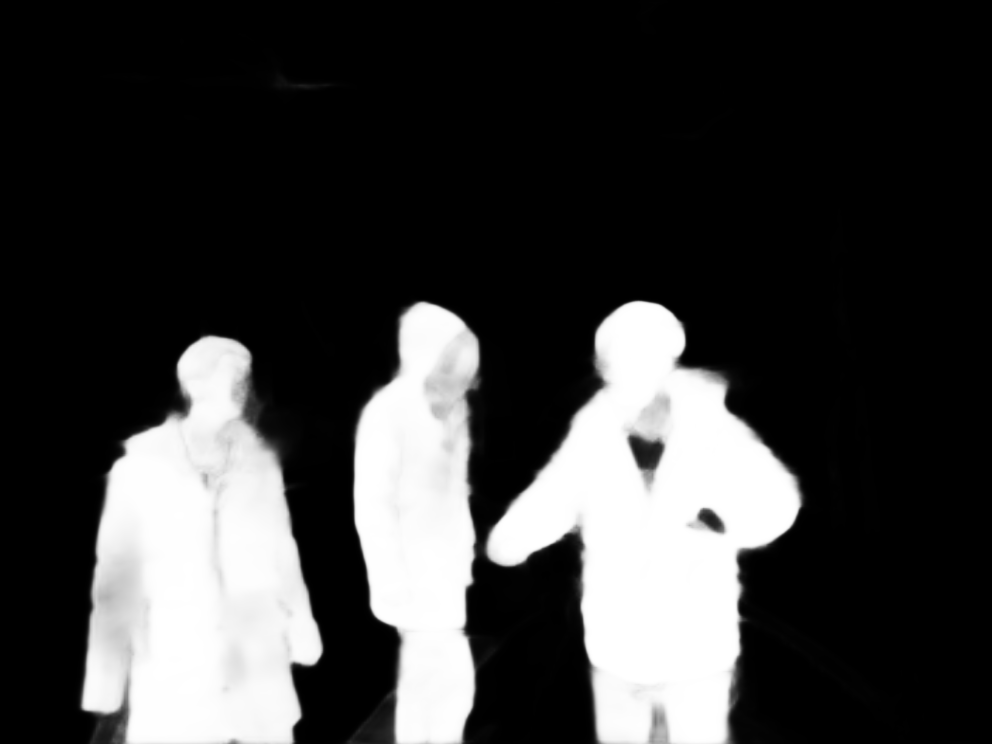}
	\end{subfigure}
	\begin{subfigure}[t]{1.13cm}
		\centering
		\includegraphics[width=1.13cm]{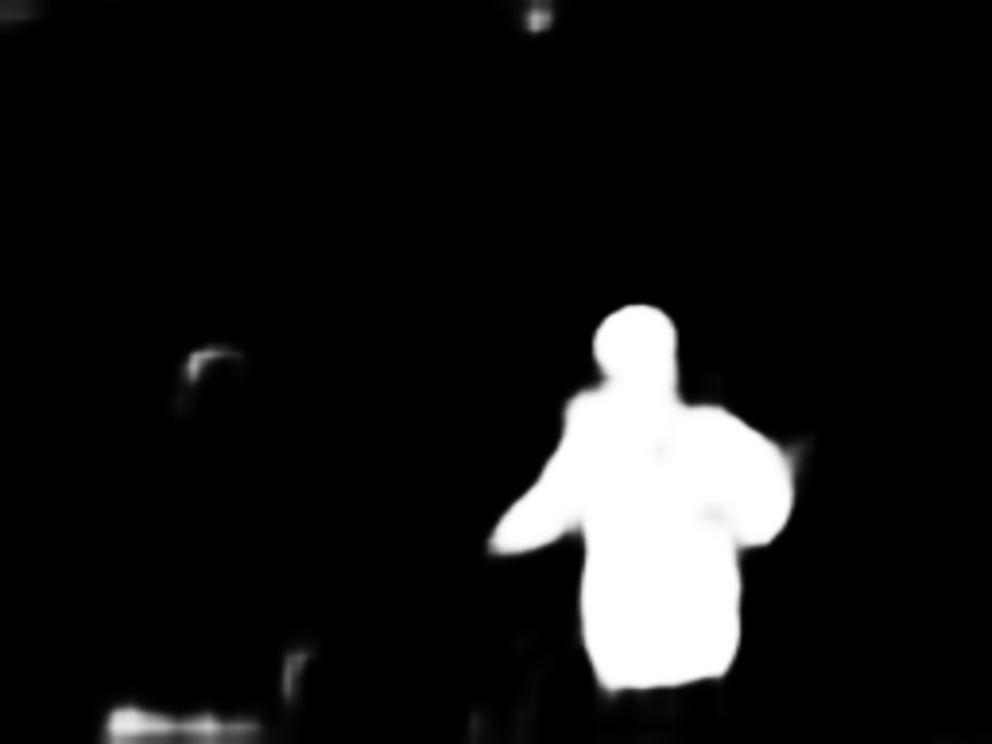}
	\end{subfigure}
	\begin{subfigure}[t]{1.13cm}
		\centering
		\includegraphics[width=1.13cm]{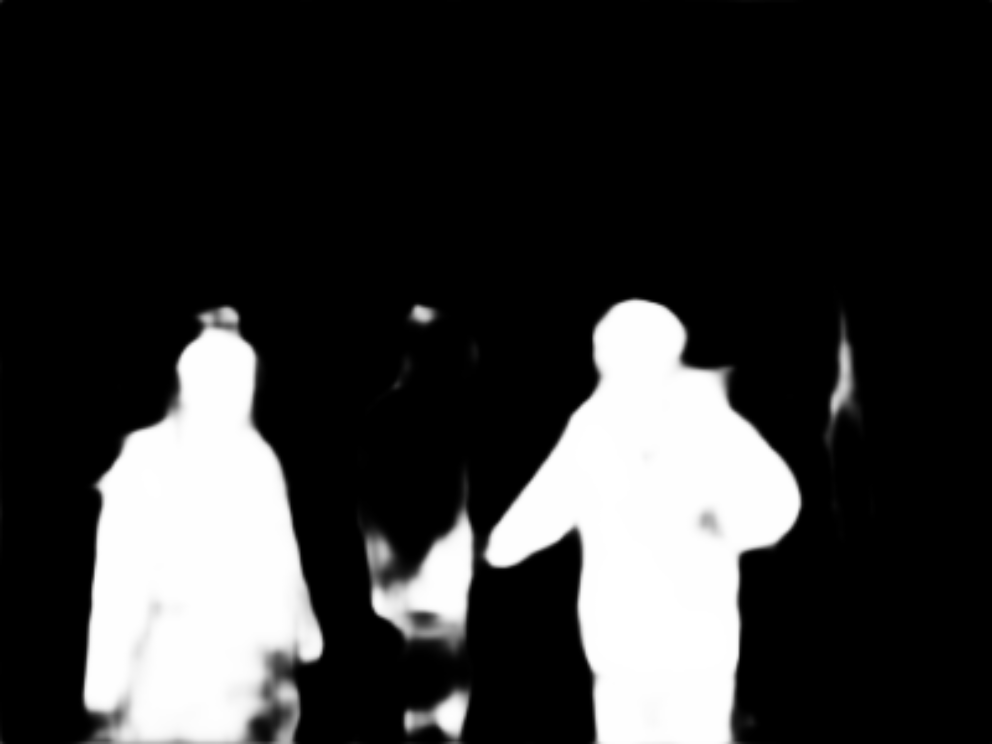}
	\end{subfigure}
	\begin{subfigure}[t]{1.13cm}
		\centering
		\includegraphics[width=1.13cm]{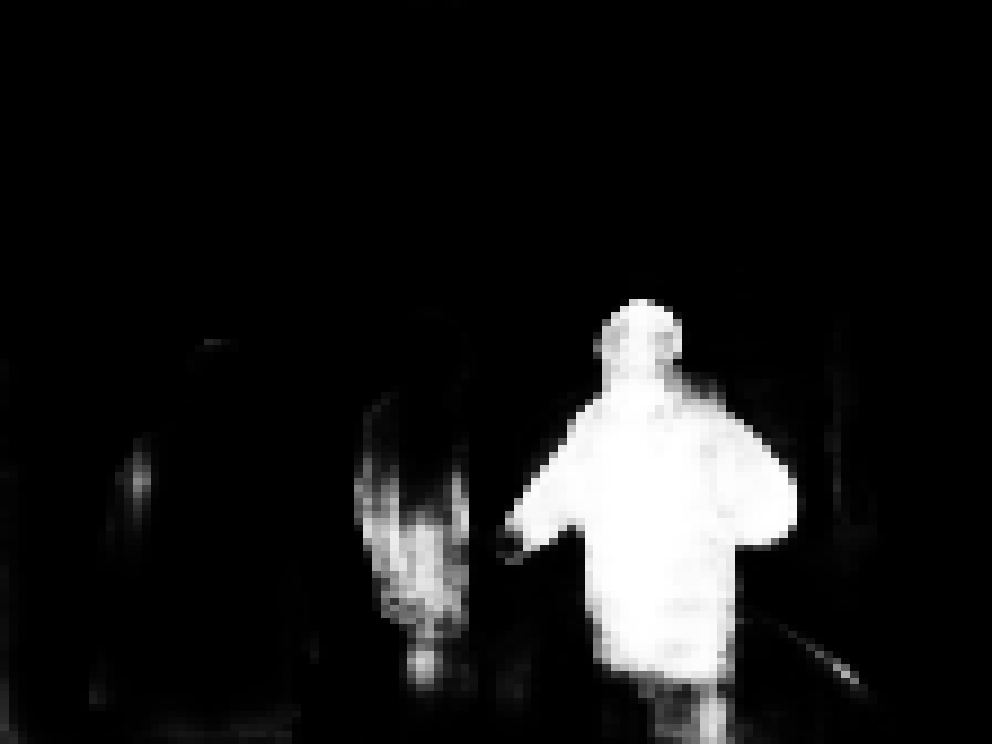}
	\end{subfigure}
	\begin{subfigure}[t]{1.13cm}
		\centering
		\includegraphics[width=1.13cm]{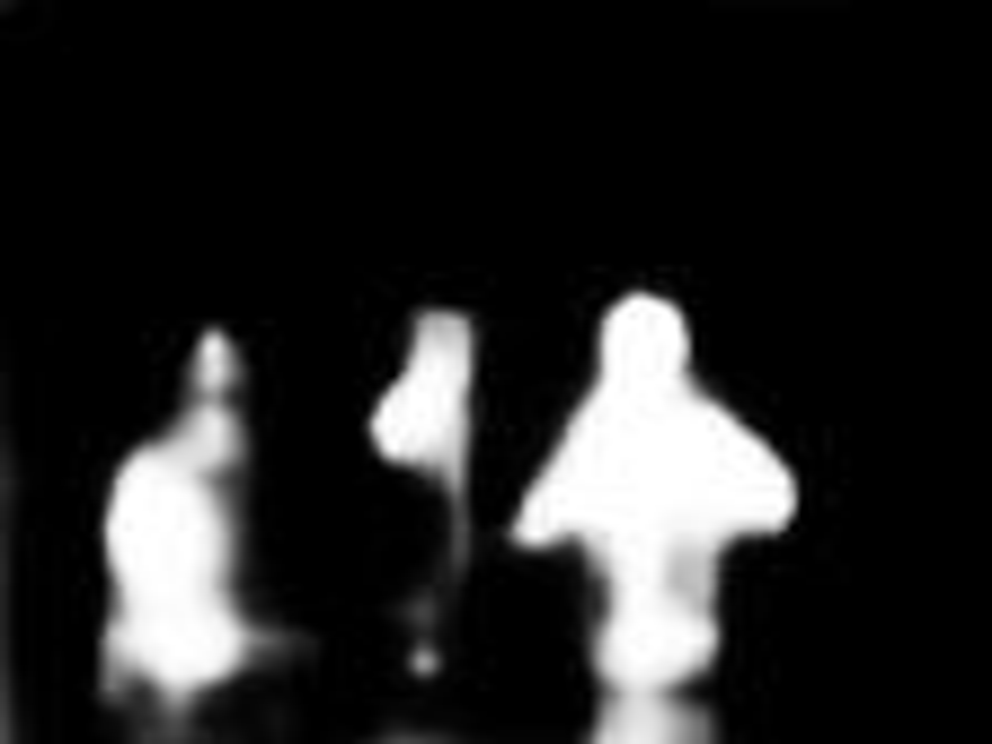}
	\end{subfigure}
	\begin{subfigure}[t]{1.13cm}
		\centering
		\includegraphics[width=1.13cm]{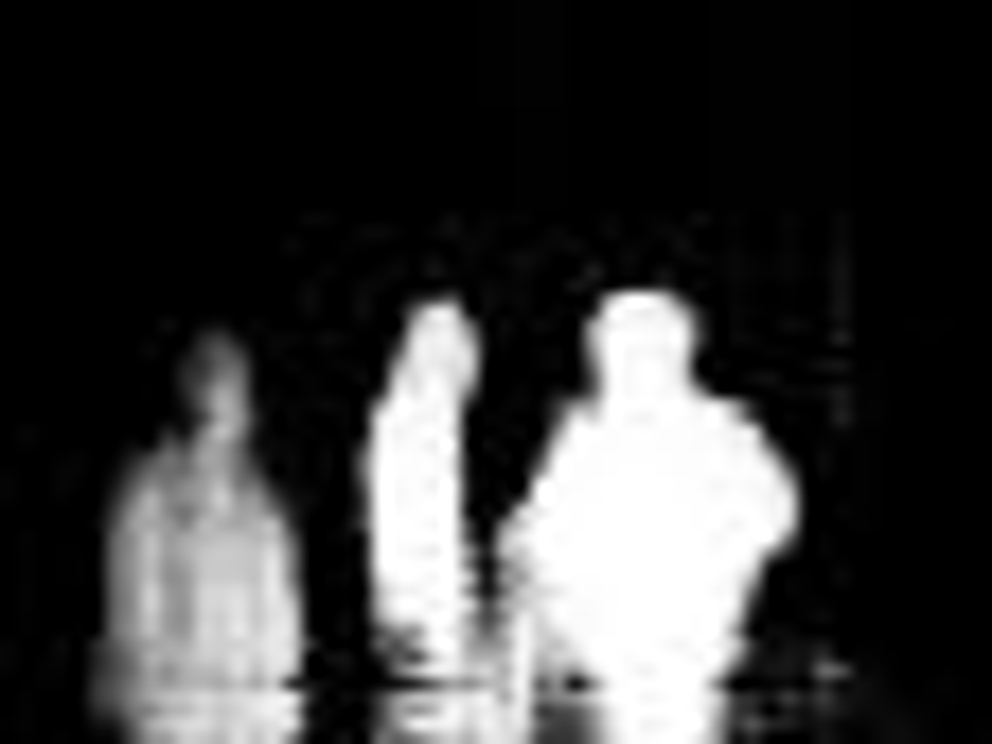}
	\end{subfigure}
	\begin{subfigure}[t]{1.13cm}
		\centering
		\includegraphics[width=1.13cm]{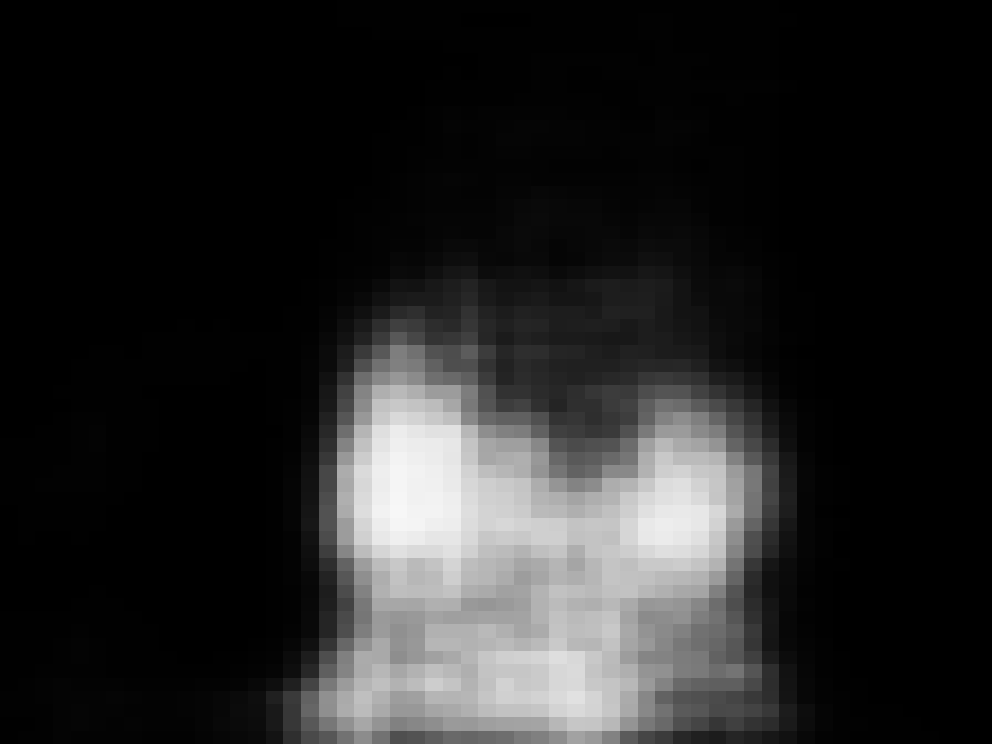}
	\end{subfigure}
	
	\vspace{1pt}

    \begin{subfigure}[t]{1.13cm}
		\centering
		\includegraphics[width=1.13cm]{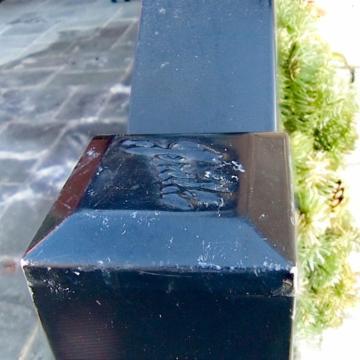}
		\caption{\scriptsize RGB}
	\end{subfigure}
	\begin{subfigure}[t]{1.13cm}
		\centering
		\includegraphics[width=1.13cm]{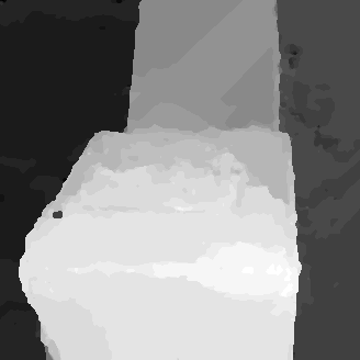}
		\caption{\scriptsize Depth}
	\end{subfigure}
	\begin{subfigure}[t]{1.13cm}
		\centering
		\includegraphics[width=1.13cm]{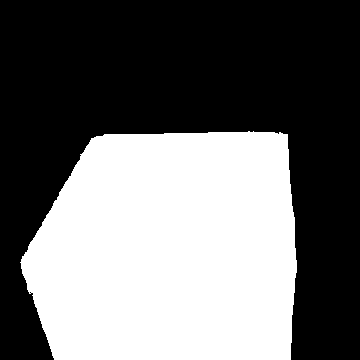}
		\caption{\scriptsize GT}
	\end{subfigure}
	\begin{subfigure}[t]{1.13cm}
		\centering
		\includegraphics[width=1.13cm]{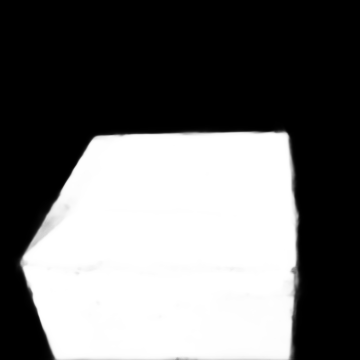}
		\caption{\scriptsize Ours*}
	\end{subfigure}
	\begin{subfigure}[t]{1.13cm}
		\centering
		\includegraphics[width=1.13cm]{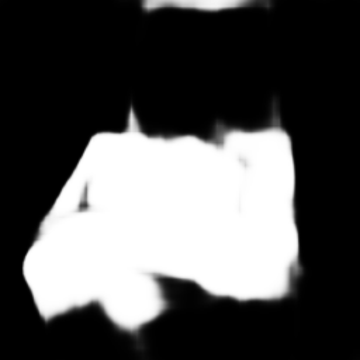}
		\caption{\scriptsize DMRA}
	\end{subfigure}
	\begin{subfigure}[t]{1.13cm}
		\centering
		\includegraphics[width=1.13cm]{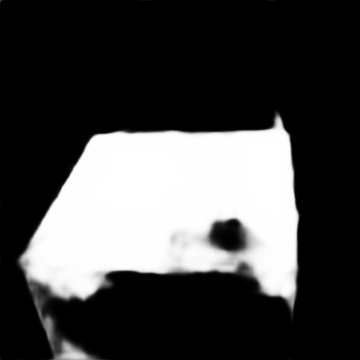}
		\caption{\scriptsize CPFP}
	\end{subfigure}
	\begin{subfigure}[t]{1.13cm}
		\centering
		\includegraphics[width=1.13cm]{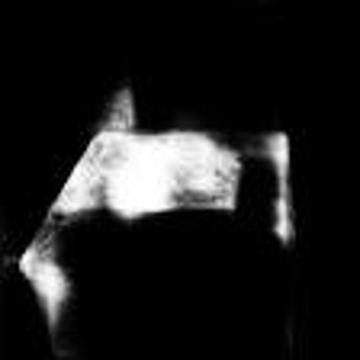}
		\caption{\scriptsize PCAN}
	\end{subfigure}
	\begin{subfigure}[t]{1.13cm}
		\centering
		\includegraphics[width=1.13cm]{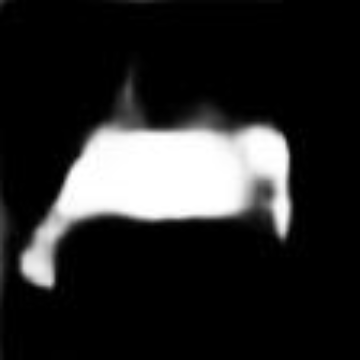}
		\caption{\scriptsize TAN}
	\end{subfigure}
	\begin{subfigure}[t]{1.13cm}
		\centering
		\includegraphics[width=1.13cm]{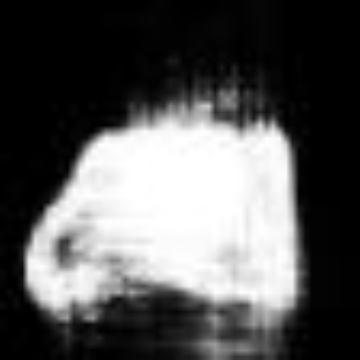}
		\caption{\scriptsize MMCI}
	\end{subfigure}
	\begin{subfigure}[t]{1.13cm}
		\centering
		\includegraphics[width=1.13cm]{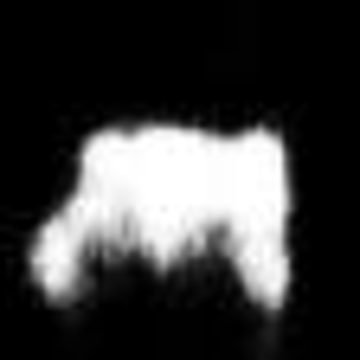}
		\caption{\scriptsize CTMF}
	\end{subfigure}
	
	\caption{Visual comparisons with state-of-the-art approaches in different challenging scenarios: low contrast in RGB or depth, complex scene, low quality depth map, multiple (small) objects, and large object. As can be seen, all the salient objects can be completely highlighted while with less false detection.}\label{fig_smaps}
\end{figure}

\textbf{Qualitative Evaluation.} We further illustrate visual examples of several representative images in different challenging scenarios to show the advantage of our method, \textit{i.e.}, low contrast in RGB or depth image (1$^{st}$-2$^{nd}$ rows), low quality depth map (3$^{rd}$-4$^{th}$ rows), complex scene (5$^{th}$ row), multiple (small) objects (6$^{th}$-7$^{th}$ rows), and large object (8$^{th}$ row). As can be seen clearly in Fig.\ref{fig_smaps} that all these cases are very challenging to the existing methods. Nevertheless, thanks to the proposed alternate refinement strategy, our model can well capture the complementary cues from the depth image, therefore, we can successfully highlight salient objects in these images, and also will not be distracted by the low quality depth maps. Furthermore, contributed by the proposed progressive guidance, the missing object parts and false detection can be well remedied, thus leads to more complete and accurate detection.

\textbf{Timing and Model Size.} The proposed network is also very efficient and compact. When trained on the NJUD+NLPR datasets with 2185$\times$4 images, our network only takes about 2.5 hours to train for 30 epochs. During the inference stage, contributed by the constructed lightweight depth stream, we can run at \textbf{71 FPS} only with \textbf{64.9 MB} model size, which is much faster and compact than the existing models as shown in Table~\ref{table_time}.

\section{Conclusions}
In this paper, we developed a progressively guided alternate refinement network for efficient RGB-D SOD. A lightweight depth stream was first constructed to extract complementary depth features by learning from scratch. Starting from a coarse initial prediction by the proposed MSR block, the RGB features and depth features are alternately fed into the designed GR blocks for progressive refinement. Contributed by the alternate refinement strategy, the mutual degradation between RGB and depth features can be well alleviated especially when the depth is low quality. With the help of the proposed progressive guidance, the missing object parts and false detection can be well refined, which resulted in more complete and accurate detection. State-of-the-art performance on 7 benchmark datasets demonstrates their effectiveness, and also shows the superiority in efficiency and compactness. In addition, the proposed network can be flexibly applied for other cross-modal SOD tasks, \textit{e.g.}, RGB-T~\cite{tang2019rgbt}. Nevertheless, the boundary details are still not accurate enough especially in high resolution images~\cite{zeng2019towards}, which will be further improved in future works. We also found that some new approaches with high performance are published after this submission, such as ICNet~\cite{li2020icnet}, UCNet~\cite{zhang2020uc}, JL-DCF~\cite{fu2020jl}, A2dele~\cite{piao2020a2dele}, and SSF~\cite{zhang2020select}. We will make a more comprehensive comparison in our extended work.

\noindent {\textbf{Acknowledgments.}} This research was supported by the National Nature Science Foundation of China (No. 61802336) and China Scholarship Council (CSC) Program. This work was mainly done when Shuhan Chen was visiting Northeastern University as a visiting scholar.

\clearpage
%
%
\bibliographystyle{splncs04}
\bibliography{pgar}
\end{document}